\documentclass{article}

\usepackage[preprint]{neurips_2025}

\usepackage[utf8]{inputenc} % allow utf-8 input
\usepackage[T1]{fontenc}    % use 8-bit T1 fonts
\usepackage[colorlinks=true, linkcolor=red, citecolor=green, urlcolor=blue]{hyperref}      % hyperlinks
\usepackage{url}            % simple URL typesetting
\usepackage{booktabs}       % professional-quality tables
\usepackage{amsfonts}       % blackboard math symbols
\usepackage{nicefrac}       % compact symbols for 1/2, etc.
\usepackage{microtype}      % microtypography
\usepackage[table]{xcolor} 
\usepackage{xspace}
\usepackage{wrapfig,booktabs,colortbl}

\usepackage{arydshln}
\usepackage{adjustbox}
\usepackage{amsmath}
\usepackage{bbding}
\usepackage{pifont}
\usepackage{comment}
\usepackage{caption}

\definecolor{custom_red}{RGB}{231,111,81}
\definecolor{custom_green}{RGB}{42,157,143}
\definecolor{custom_dark}{RGB}{38,70,83}
\definecolor{custom_yellow}{RGB}{233,196,106}
\definecolor{custom_orange}{RGB}{244,162,97}

\newcommand{\methodname}{4Real-Video-V2\xspace}

\title{4Real-Video-V2: Fused View-Time Attention and Feedforward Reconstruction for 4D Scene Generation}
\author{%
  Chaoyang Wang\text{$^{1}$}\thanks{Equal contribution.} \quad Ashkan Mirzaei \text{$^{1}$}\footnotemark[1] \quad
  Vidit Goel\text{$^{1}$} \quad
  Willi Menapace\text{$^{1}$} \quad
  Aliaksandr Siarohin\text{$^{1}$}  \\
  \textbf{Avalon Vinella\text{$^{1}$}} \quad
  \textbf{Michael Vasilkovsky\text{$^{1}$}} \quad
  \textbf{Ivan Skorokhodov\text{$^{1}$}} \quad
  \textbf{Vladislav Shakhrai\text{$^{1}$}} \\
  \textbf{Sergey Korolev\text{$^{1}$}} \quad
  \textbf{Sergey Tulyakov\text{$^{1}$}} \quad
  \textbf{Peter Wonka\text{$^{1,2}$}} \quad
  \\
  \text{$^{1}$}Snap Inc. \quad
  \text{$^{2}$}KAUST\\
  {\href{https://snap-research.github.io/4Real-Video-V2/}{Project page: https://snap-research.github.io/4Real-Video-V2/}}
}

\begin{document}

\maketitle

\begin{figure}[htbp]
    \vspace{-6mm}
    \makebox[\linewidth][c]{%
        \includegraphics[width=1\linewidth,trim={0.0cm 0 0 0},clip]{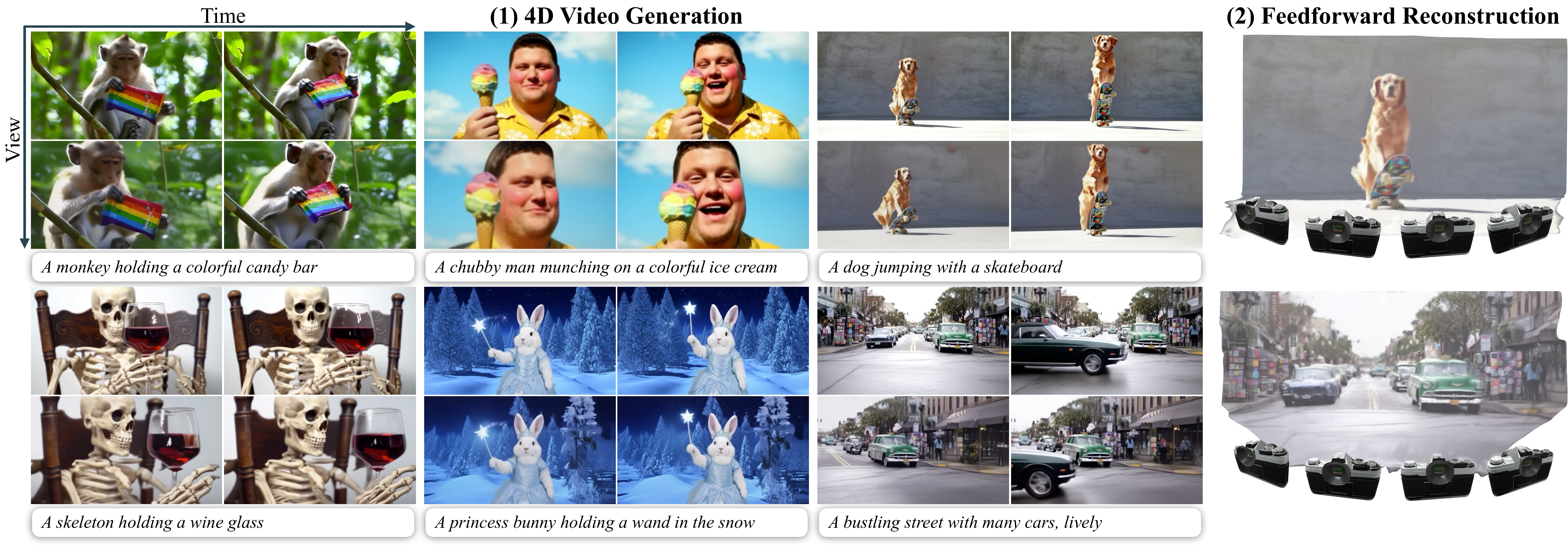}
    }
    \vspace{-6mm}
    \caption{Our method enables the creation of 4D scenes from a text prompt by combining a diffusion model that directly generates synchronized multi-view videos with a feedforward reconstruction model that efficiently produces Gaussian-based representations.}
    \label{fig:teaser}
\end{figure}

\begin{abstract}
  % We propose a first framework that can compute a 4D spatio-temporal grid of video frames and 3D Gaussian point clouds for each time step in a feed-forward architecture. In the first part of our paper, we provide an analysis for the current 4D video diffusion architectures that perform spatial and temporal attention either subsequently or in parallel in a two stream architecture. We highlight the limitations of existing work and propose a novel fused architecture that can perform spatial and temporal attention in a single layer. The key to our work is a sparse attention pattern, in which tokens attend to other tokens in the same frame, other tokens that share the same timestamp, and other tokens that share the same viewpoint. In the second part, we describe an extension to existing 3D reconstruction algorithms, by introducing a Gaussian head, a camera token replacement algorithm, and additional dynamic training.
% Overall, we set the new SOTA for 4D video and improve upon the current state of the art in both visual quality as well as reconstruction capability.
\vspace{-3mm}
We propose the first framework capable of computing a 4D spatio-temporal grid of video frames and 3D Gaussian particles for each time step using a feed-forward architecture. Our architecture has two main components, a 4D video model and a 4D reconstruction model. In the first part, we analyze current 4D video diffusion architectures that perform spatial and temporal attention either sequentially or in parallel within a two-stream design. We highlight the limitations of existing approaches and introduce a novel fused architecture that performs spatial and temporal attention within a single layer. The key to our method is a sparse attention pattern, where tokens attend to others in the same frame, at the same timestamp, or from the same viewpoint.
In the second part, we extend existing 3D reconstruction algorithms by introducing a Gaussian head, a camera token replacement algorithm, and additional dynamic layers and training. Overall, we establish a new state of the art for 4D generation, improving both visual quality and reconstruction capability.

\vspace{-4mm}
\end{abstract}

\section{Introduction}
\vspace{-2mm}

\begin{table}[ht]
\caption{Comparison of recent 4D video generation methods. We use \emph{Modified} to denote methods that rely on sophisticated adjustments to the diffusion process for generating 4D videos. A question mark (?) indicates methods that are theoretically extendable to 4D video generation, but such extensions were not explored in their original papers.}
    \label{tab:4d_video_categories}
    \centering
    \scriptsize
    \resizebox{\textwidth}{!}{
    \begin{tabular}{cllllcc}
    \toprule
     Category     & Method          & Model Output & Scene Type & 4D Inference & DiT-based & Temporal Compress \\
    \midrule
         & 4DiM~\cite{4DiM} & Images & Scene & ? & \textcolor{custom_red}{\ding{55}} & \textcolor{custom_red}{\ding{55}} \\
    4D-Aware     & CAT4D~\cite{cat4d} & Images & Scene &   Modified & \textcolor{custom_red}{\ding{55}} & \textcolor{custom_red}{\ding{55}} \\
    Image/Video Gen     & GenXD~\cite{genxd} & Video & Scene \& Object & Modified & \textcolor{custom_red}{\ding{55}} & \textcolor{custom_red}{\ding{55}} \\
         & DimensionX~\cite{sun2024dimensionx} & Video & Scene & Modified & \textcolor{custom_green}{\ding{51}} & \textcolor{custom_green}{\ding{51}} \\
    \hdashline
    3D Point Cloud    & GEN3C~\cite{gen3c} & Video & Scene & ? & \textcolor{custom_green}{\ding{51}} & \textcolor{custom_green}{\ding{51}} \\
    Cond. Video Gen    & TrajectoryCrafter~\cite{yu2025trajectorycrafterredirectingcameratrajectory} & Video & Scene & ? & \textcolor{custom_green}{\ding{51}} & \textcolor{custom_green}{\ding{51}} \\
    \hdashline
    Video-Video & Generative Camera Dolly~\cite{van2024generative} & Video & Scene & ? & 
    \textcolor{custom_red}{\ding{55}} & \textcolor{custom_red}{\ding{55}} \\
    Gen & ReCamMaster~\cite{recammaster} & Video & Scene & ? & 
    \textcolor{custom_green}{\ding{51}} & \textcolor{custom_green}{\ding{51}} \\
    \hdashline
    & CVD~\cite{cvd} & MV Video & Scene & Native (2 View) & \textcolor{custom_red}{\ding{55}} & \textcolor{custom_red}{\ding{55}}\\
    & Human4DiT~\cite{human4dit} & MV Video & Human & Native & \textcolor{custom_green}{\ding{51}} & \textcolor{custom_red}{\ding{55}}\\
    & VividZoo~\cite{vividzoo} & MV Video & Object & Native & \textcolor{custom_red}{\ding{55}} & \textcolor{custom_red}{\ding{55}}\\
    4D (Sychronized & 4Diffusion~\cite{4diffusion} & MV Video & Object & Native & \textcolor{custom_red}{\ding{55}} & \textcolor{custom_red}{\ding{55}}\\
    Multi-View) Video Gen & SV4D~\cite{sv4d} & MV Video & Object & Native & \textcolor{custom_red}{\ding{55}} & \textcolor{custom_red}{\ding{55}}\\
    & SynCamMaster~\cite{syncammaster} & MV Video & Scene & Native & \textcolor{custom_green}{\ding{51}} & \textcolor{custom_green}{\ding{51}}\\
    & 4Real-Video~\cite{4real-video} & MV Video & Scene & Native & \textcolor{custom_green}{\ding{51}} & \textcolor{custom_red}{\ding{55}}\\
    & Ours & MV Video & Scene \& Object & Native & \textcolor{custom_green}{\ding{51}} & \textcolor{custom_green}{\ding{51}} \\
    \bottomrule
    \end{tabular}
    }
    \vspace{-5mm}
\end{table}

Immersive visual experiences are becoming increasingly popular in fields such as virtual reality and film production. This growing demand drives the need for technologies that enable the creation of 4D content, where users can interactively explore dynamic scenes.
A key challenge in enabling such capabilities is the limited availability of high-quality 3D and 4D data. This scarcity presents a significant obstacle to training generative models that can directly produce 4D representations.

In contrast, video generation has made rapid progress in recent years~\cite{cogvideox,wan2025,stepfun,moviegen,veo2}, driven by large-scale datasets and advances in diffusion models. Building on this progress, several recent works~\cite{4real-video,cat4d,sv4d,4diffusion} extend video generation to the 4D domain by producing what we refer to as 4D videos. These are synchronized multi-view video grids that can supervise reconstruction methods to recover explicit 4D representations such as dynamic NeRFs~\cite{nerf} or Gaussian splats~\cite{kerbl3Dgaussians}.
We also adopt this two-stage approach because it leverages the strong priors of pretrained video models and offers a promising path toward generalizable and photorealistic 4D generation.

In the first stage, as categorized in Tab.~\ref{tab:4d_video_categories}, some prior methods attempt to enhance 2D video models with camera and motion control~\cite{cat4d,genxd,sun2024dimensionx,4DiM}, 3D point cloud condioning~\cite{yu2025trajectorycrafterredirectingcameratrajectory,gen3c}, or reference-video conditioning~\cite{recammaster,van2024generative}. However, these models do not natively generate synchronized multi-view outputs which often results in degraded quality and reduced efficiency.
% typically rely on inference-time strategies such as alternating diffusion steps across time and views. 
% These strategies often result in degraded quality and reduced efficiency.

Other approaches attempt to directly generate the complete multi-view video grid, where each row corresponds to a specific freeze-time step and each column to a fixed viewpoint. Most of these methods~\cite{sv4d,human4dit,vividzoo,4diffusion} are designed for object-centric data and do not generalize to complex scenes. Among the few general-purpose models, SynCamMaster~\cite{syncammaster} is trained to generate a sparse set of diverse viewpoints, but often suffers from poor consistency across views. 4Real-Video~\cite{4real-video} achieves stronger multi-view alignment by training with dense viewpoints, but is only tested on a smaller architecture that lacks features of modern video models such as temporal compression and higher resolution.
% its pixel-based architecture, lacking temporal compression, is constrained to low resolution, inefficient for long videos and struggles to render high-frequency details in complex environments.

% to do: shorten the discussion of other works.
% to do: establish the problem, analysis other architecture
When scaling to modern architectures, it is important to consider their large number of parameters (e.g., we use an 11B-parameter base model). Given the limited availability of 4D video training data and the significantly increased number of tokens introduced by handling multiple viewpoints, the design of a 4D video model becomes critically important. The two main current architectures are the sequential architecture interleaving spatial and temporal attention~\cite{vividzoo,syncammaster}, and the parallel architecture that computes spatial and temporal attention in parallel and merges the results~\cite{4real-video}. By experimenting with multiple architecture variations, we believe that the most important criteria for developing an architecture is to minimize the number of new parameters that need to be trained from scratch and to minimize fine-tuning, so that each layer is used similarly to its pretraining.
% To overcome these limitations, we introduce a new model architecture for 4D video generation that is compatible with most modern transformer-based video models. 
Building on this analysis, our key contribution is a parameter-efficient design that introduces no additional parameters to the base model. Specifically, we fuse cross-view and cross-time attention into a single self-attention mechanism. In contrast to previous approaches that apply these attentions separately and introduce new attention~\cite{syncammaster,human4dit,4diffusion,vividzoo} or synchronization modules~\cite{4real-video}, our unified formulation allows us to take advantage of highly optimized sparse attention implementations. This results in minimal computational overhead and enables effective fine-tuning of large pretrained video models.

In the second stage, traditional reconstruction methods often rely on iterative optimization to recover 4D representations from multi-view video inputs. Although accurate, these methods tend to be slow, sensitive to camera estimation errors, and difficult to scale to dynamic scenes of longer duration. To address this, we extend a state-of-the-art feedforward 3D reconstruction~\cite{wang2025vggt} to directly predict both camera parameters and time-varying Gaussian splats from synchronized multi-view video frames. This approach greatly improves efficiency while preserving visual quality.

In summary, our contributions are:
1)
A novel two-stage 4D generation framework that produces a grid of images and converts them into Gaussian ellipsoids.
2)
A fused view and time attention mechanism that enables parameter-efficient 4D video generation.
3)
A feedforward model that jointly recovers camera parameters and Gaussian particles from multi-view videos.

\section{Related Work}
\vspace{-3mm}

\paragraph{Optimization-based 4D generation.}
Score Distillation Sampling (SDS)\cite{dreamfusion,prolificdreamer,sjc,fantasia3d,magic3d,hifa} is a common method for creating 3D scenes. It uses gradients from pre-trained models like text-to-image~\cite{ldm,imagen} and text-to-multi-view~\cite{zero123,mvdream} models. Recent 4D methods~\cite{4dfy,ayg,consistent4d,dreamgaussian4d,4dgen,animate124,mav3d,4real} extend this by using text-to-video models~\cite{animatediff,imagenvideo,videocrafter} to add motion. These methods usually take hours to run because they rely on slow optimization. Most of them also use 3D priors from diffusion models~\cite{zero123,mvdream} trained on synthetic object-centric datasets like Objaverse~\cite{objaverse}, which can make the results look unrealistic and limited to single objects.

\noindent\textbf{Camera-Aware video generation.}
Text-to-video models~\cite{sora,cogvideox,moviegen,veo2,snapvideo} have made significant progress in generating realistic videos. To provide users with more control, some methods incorporate camera motion by leveraging camera pose data~\cite{vd3d,motionctrl,cameractrl,directavideo,4DiM}, while others fine-tune models using videos annotated with camera labels~\cite{genxd,camco,sun2024dimensionx}. 
CVD~\cite{cvd} and ReCamMaster~\cite{recammaster} further enable modifying camera motion of on existing footage.
Another line of work introduces a 3D cache, such as a point cloud, to store scene geometry. These representations are then projected into novel views and completed using diffusion models~\cite{yu2025trajectorycrafterredirectingcameratrajectory,gen3c,wang2024freevs}. These camera-aware techniques enable impressive 3D visual effects, such as dolly shots and bullet-time sequences. However, these methods do not natively generate a complete set of frames across both time and viewpoints. Extending them to 4D video requires substantial modifications to the diffusion process~\cite{cat4d,sun2024dimensionx}, often leading to artifacts and quality degradation.

\noindent\textbf{4D video generation.}
We define 4D video (or synchronized multi-view video) as a grid of video frames organized along both temporal and viewpoint dimensions. Several methods~\cite{sv4d,vividzoo,4diffusion,human4dit} are trained on 4D datasets derived from Objaverse~\cite{objaverse} or human motion capture sequences. While these models aim to generalize beyond single-object scenes, they are still limited by the lack of diverse and scalable 4D training data. CVD~\cite{cvd} addresses this limitation by fine-tuning models to generate synchronized video pairs using pseudo-paired samples from real-world datasets~\cite{webvid10m,realestate10k}. SynCamMaster~\cite{syncammaster} and 4Real-Video~\cite{4real-video} further extend this direction by generating synchronized multi-view videos using a combination of synthetic 4D and real 2D datasets. SynCamMaster is trained on sparsely sampled viewpoints, allowing for good view control but showing inconsistencies across views. 4Real-Video, on the other hand, uses densely sampled, continuous camera trajectories to improve view consistency. However, it is built on a relatively small pixel-based video backbone, which limits visual quality and lacks temporal compression, making it inefficient for longer videos. Our work improves upon 4Real-Video by introducing a more efficient architecture that scales effectively with large video generation models.

\noindent \textbf{Feed-forward reconstruction.} Recent advances in 3D reconstruction increasingly use data-driven priors to speed up the process~\cite{fan2024instantsplat}. Some methods use priors for guidance or initialization to enable faster optimization~\cite{Jain_2021_ICCV,chen2024g3r,23iccv_tian_mononerf}, but still rely on few-shot optimization to refine results~\cite{gntmove2023,t2023is}. To avoid optimization, newer work explores fully feedforward models that infer 3D scenes from 2D inputs. These models often focus on static scenes and represent geometry using triplanes~\cite{hong2023lrm}, 3D Gaussians~\cite{charatan23pixelsplat,chen2024mvsplat,chen2024mvsplat360,gslrm2024}, sparse voxel grids~\cite{ren2024scube}, or learned tokens~\cite{jin2025lvsm,jiang2025rayzer}. They usually need ground-truth camera calibration or rely on traditional methods to estimate camera parameters, which can be slow and unreliable for generated assets. This has driven interest in pose-free, feedforward reconstruction for static scenes~\cite{smart2024splatt3r,zhang2025FLARE,ye2024noposplat,hong2024pf3plat}. While these models work well for static scenes, handling dynamic scenes is still a challenge. Current dynamic scene methods often assume dense, temporally consistent video depth maps~\cite{zhao2024pgdvs}, which are hard to obtain. Others lack rendering support~\cite{zhang2024monst3r,Wang_2024_CVPR,mast3r_arxiv24,wang2025vggt}, or only work with object-centric data~\cite{ren2024l4gm}. Some also assume known camera poses and monocular video~\cite{liang2024btimer}. Another challenge is that even when RGB loss is used for realistic outputs, many methods produce poor geometry, limiting novel view synthesis to small camera movements~\cite{gslrm2024,liang2024btimer}.

\begin{figure}
    \centering
    \includegraphics[width=\linewidth]{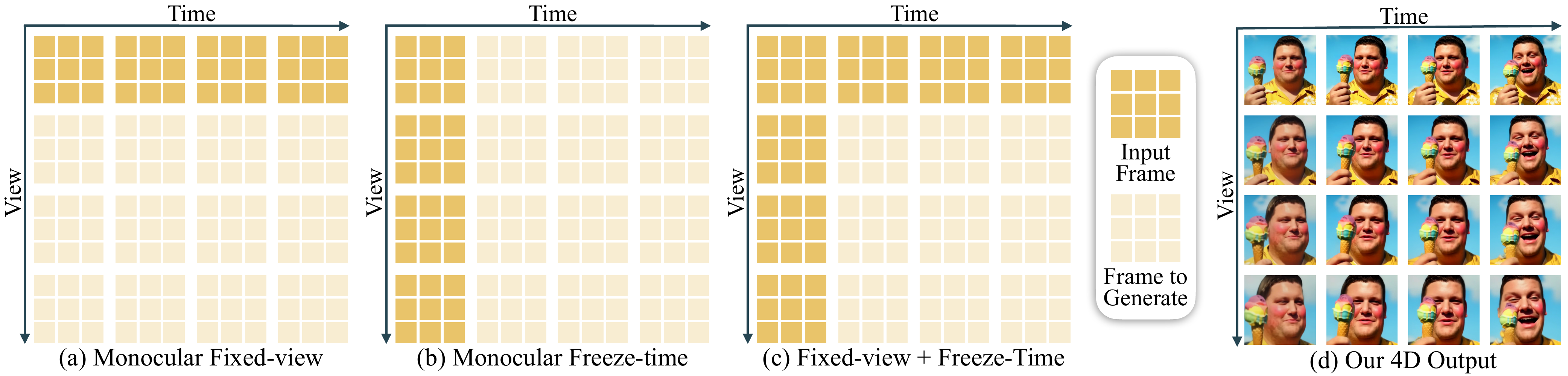}
    \vspace{-3mm}
    \caption{Our 4D video model supports input types including: (a) a fixed-view video, (b) a freeze-time video showing multiple angles of a scene at a single timestep, and (c) a combination of both. Each input can be generated from a text prompt using standard video models. }
    \label{fig:4d_video_inputs}
    \vspace{-4mm}
\end{figure}
\begin{figure}[h]
    \centering
    \includegraphics[width=0.95\linewidth]{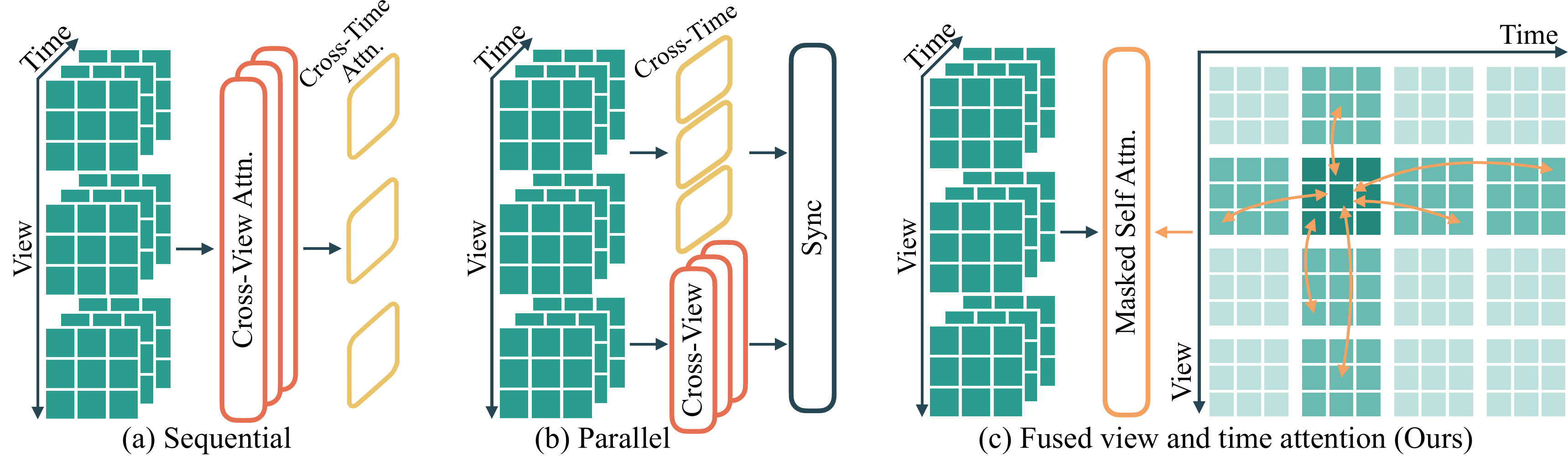}
    \caption{We analyze three architectures for 4D video generation: (a) sequential cross-view and cross-time attention, (b) parallel cross-view and cross-time attention with an extra synchronization layer, and (c) our proposed architecture using fused view-time attention with masked self-attention.}
    \label{fig:prelim_architecture}
    \vspace{-4mm}
\end{figure}
\vspace{-3mm}
\section{Method}
\vspace{-3mm}
Our 4D generation pipeline comprises two main stages. First, we introduce a novel 4D video diffusion model that generates synchronized multi-view videos of dynamic scenes across time and viewpoints (Sec.~\ref{subsec:MVVideo}). Next, we apply a purpose-built feedforward reconstruction network to lift Gaussian ellipsoids from the generated frames (Sec.~\ref{subsec:FFReconstruction}). We detail each component below.
\vspace{-2mm}
\subsection{Synchronized Multi-view Video Diffusion}
\label{subsec:MVVideo}
\vspace{-2mm}
We aim to generate a structured grid of video frames $\{I_{v,t}\}$, where all frames in a row share a viewpoint $v$, and all frames in a column share a timestep $t$. In other words, each row is a fixed-view video, and each column is a freeze-time video.

\noindent\textbf{Preliminary I: DIT-based Diffusion Transformer.} The Diffusion Transformer (DiT)~\cite{Peebles2022DiT} architecture has been widely adopted in modern video diffusion models~\cite{cogvideox,wan2025,cosmos,moviegen,stepfun} for denoising latent video tokens. The model takes as input a set of latent tokens $\{ \mathbf{x}_{t,x,y} \in \mathbb{R}^d \}$, where $t$, $x$, and $y$ index the temporal and spatial dimensions, and $d$ is the latent dimension. These tokens represent compressed versions of high-resolution videos, typically downsampled by a factor of $8\times$ spatially and $4\times$ or $8\times$ temporally. The tokens are first embedded via a patch embedding layer and then processed through a series of DiT blocks. Each block contains a 3D self-attention layer that jointly attends the features across both spatial and temporal dimensions, followed by a cross-attention layer that conditions the features on input context such as text embeddings.

\textbf{Preliminary II: Prior architectures for synchronized multi-view video diffusion.} Standard video diffusion models generate a single sequence of frames with entangled view changes and scene motion. To extend pretrained video diffusion models for generating synchronized multi-view videos, prior works introduce additional cross-view self-attention layers to enforce consistency across views. As illustrated in Fig.~\ref{fig:prelim_architecture},  these methods can be categorized based on how the cross-view attention is incorporated into the architecture.

\emph{Sequential architecture.} These methods~\cite{vividzoo, syncammaster, human4dit, 4diffusion, sv4d} \emph{sequentially} interleave cross-view and cross-time self-attention layers. The cross-view layers are typically initialized from multi-view image models trained to synthesize static scenes from different viewpoints, while the cross-time layers are initialized from pretrained video models. These attention layers are either jointly trained~\cite{human4dit, sv4d}, or fine-tuned selectively by freezing one type while updating the other~\cite{syncammaster, 4diffusion}. 

\emph{Parallel architecture.} An alternative strategy applies cross-view and cross-time self-attention in parallel, rather than interleaving them sequentially~\cite{4real-video, sv4d-v2}. In this setup, each attention branch processes the video tokens independently—cross-view attention enforces spatial consistency across viewpoints, while cross-time attention captures temporal dynamics. The outputs from both branches are then fused through a synchronization module designed to align and integrate the two branches. This decoupled design has two key advantages: (1) it avoids interference between the attention branches, which are originally trained to operate in isolation, and (2) it enables reusing frozen pretrained models for both branches, reducing training cost and preserving their generalization capability. Only the lightweight synchronization module needs to be trained~\cite{4real-video}.

\noindent\textbf{Fused view and time attention architecture.} Parallel architectures introduce additional parameters through the synchronization module. To mitigate overfitting on limited 4D data, these modules are deliberately kept lightweight, typically implemented as a single linear layer~\cite{4real-video} or a weighted averaging operation~\cite{sv4d-v2}. In contrast, we propose a new design that requires no additional parameters beyond a pretrained video model, and as a result, it naturally maintains strong generalization without relying on manually crafted bottleneck layers.

Specifically, we propose to fuse cross-view and cross-time attention into a single self-attention operation. For each latent token $\mathbf{x}_{v,t,x,y}$ representing view $v$, time $t$, and spatial location $(x,y)$, features are computed by attending to all other tokens sharing either the same view or timestamp. This is implemented using a masked self-attention layer that enforces the desired attention pattern:
\begin{equation}
\vspace{-1mm}
\resizebox{0.94\textwidth}{!}{\ensuremath{
  \text{SoftMax}( \mathbf{M} \odot 
  \frac{\mathbf{\mathbf{Q}\mathbf{K}^T}}{\sqrt{d}}) \mathbf{V}, \quad \mathbf{M}(\text{Idx}(v_q,t_q,x_q,y_q), \text{Idx}(v_k,t_k,x_k,y_k)) = \begin{cases}
1, & \text{if}~v_q = v_k, ~\text{or}~t_q = t_k \\
0, & \text{otherwise}
\end{cases}
}}
\end{equation}
where $\mathbf{Q}$, $\mathbf{K}$, $\mathbf{V} \in \mathbb{R}^{N \times d}$ are the queries, keys and values of all tokens, $N=VTHW$ is the total number of tokens, $\odot$ denotes element-wise multiplication, and $\mathbf{M}\in\mathbb{R}^{N \times N}$ is a binary mask. The function ${\text{Idx}()}$ maps the view, temporal and spatial indices of a token to its corresponding flattened index.

The masked self-attention is efficiently implemented using FlexAttention~\cite{flexattention}, which exploits the sparsity of the attention mask to reduce memory and computation. With a high sparsity ratio of $1 - \frac{T+V}{TV}$, the approach scales efficiently to a large number of views and timestamps.

\noindent\textbf{Positional embedding.} For standard 2D video generation, each token is associated with a 3D positional embedding $(t, x, y)$. In contrast, multi-view videos introduce an additional view dimension, resulting in 4D indices $(v, t, x, y)$. Directly using 4D positional embeddings would introduce significant discrepancies with the pretrained video model. To address this, we map the 4D coordinates to 3D by collapsing the view and time dimensions: $(v,t,x,y)\rightarrow(v*T_{\text{max}}+t, x, y)$, where $T_\text{max}$ is the maximum temporal length supported by the model. This is equivalent to flatten the multi-view video into a single long sequence. A similar idea was independently adopted by ReCamMaster~\cite{recammaster} for the specific case of synchronizing two videos. We also explored alternative transformations (see Supplementary), but found the above approach to be the most effective.

\noindent\textbf{Temporally compressed latents.}
Each latent tokens compresses patches from 4 consequtive frames of the same view points. We choose to compress along the temporal dimension rather than the view dimension, as densely sampled viewpoints offer limited benefit to reconstruction quality in the second stage. In contrast, temporal compression significantly improves generation throughput.

\begin{figure}
\centering
\includegraphics[width=0.9\linewidth]{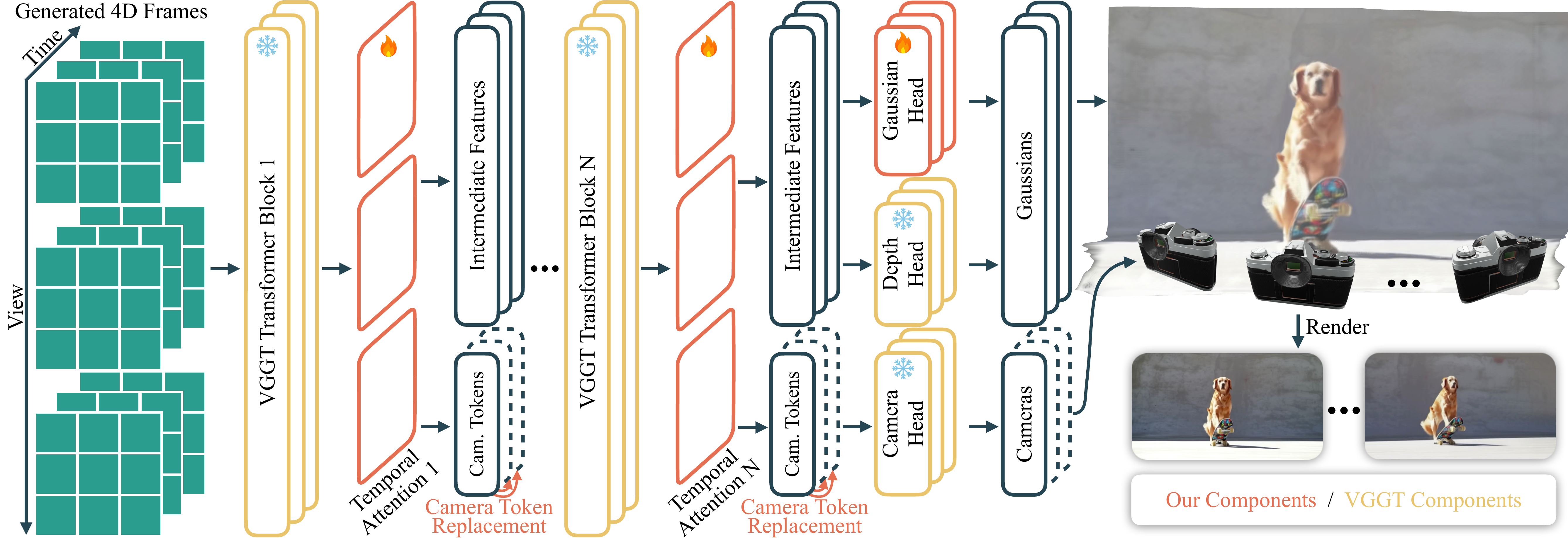}
\vspace{-1mm}
\caption{Overview of our feedforward reconstruction model. Built on top of \textcolor{custom_yellow}{VGGT}~\cite{wang2025vggt}, it incorporates \textcolor{custom_red}{temporal attention} layers, \textcolor{custom_red}{camera token replacement} to ensure consistent cameras over time, and a \textcolor{custom_red}{Gaussian head} that predicts Gaussian parameters.}
\label{fig:feedforward_reconstruction}
\vspace{-4mm}
\end{figure}

\noindent\textbf{Grounded generation with reference videos.} Inspired by prior works~\cite{4real-video,sv4d}, we enable the model to condition on two types of reference videos: a \emph{fixed-view} video, which specifies the content and object motion of the dynamic scene from a single viewpoint, and a \emph{freeze-time} video, which captures the scene from multiple views at a single timestamp (see illustration in Fig.~\ref{fig:4d_video_inputs}). The model then generates all other frames corresponding to unseen view-time combinations. The conditioning is implemented by adding the patch embeddings of the reference frames to the noised latent tokens before feeding them into the denoising transformer. Unlike prior works that rely on explicit camera pose embeddings, our model learns to infer viewpoints directly from the input reference videos. 

% This design enhances both generation quality and application flexibility, as it allows the use of diverse input sources—either videos generated by state-of-the-art models or real-world captured footage—and supports applications such as re-viewing a scene from new angles or animating a 3D scene from static observations. 
%This design enhances both generation quality and application flexibility, as it allows the use of diverse input sources, such as videos generated by state-of-the-art models or real-world captured footage, and supports applications like re-viewing a scene from new angles or animating a 3D scene from static observations.
% This design choice improves usability for 3D scene animation tasks, as it avoids the need for explicit pose estimation and scale alignment—steps that are often technically challenging for non-expert users.
This design choice improves usability for 3D scene animation tasks, as it avoids the need for explicit pose estimation and scale alignment, which are often technically challenging for non-expert users.

\noindent\textbf{Training.} The model is trained using the same rectified flow objective~\cite{liu2022flow} as the base video model, following a progressive schedule that gradually increases the temporal duration. Only the self-attention layers are fine-tuned, while the rest of the model remains frozen. Training is efficient, and we report results after 4k iterations with a batch size of 96. Additional model details are provided in the supplementary. The training data includes:
1) \textit{Synthetic multi-view videos}, rendered using animated 3D assets~\cite{objaverse} and physics-based simulations~\cite{kubric}.
% 2) \textit{2D transformed videos}, generated by applying randomly sampled 2D homographic transformations to each frame of a video to simulate synchronized multi-view captures. This type of data serves as an essential data augmentation to limited real 4D data. 
2) \textit{2D transformed videos}, created by applying random 2D homographic transformations to each video frame to simulate synchronized multi-view captures. This data is a key augmentation for the limited real 4D data.
3) \textit{3D videos}, depicting static scenes or objects recorded along continuous camera trajectories. These are temporally duplicated to simulate multi-view sequences without dynamic motion.

\begin{table}[]
\caption{Quantitative comparison of our architecture (fused view-time attention) with baselines.}
\label{tab:objaverse_and_nvidia}
    \centering
    \scriptsize
    \begin{tabular}{lcccccc}
    \toprule
               & \multicolumn{3}{c}{Objaverse} & \multicolumn{3}{c}{NVIDIA Dynamic Dataset~\cite{yoon2020novel}} \\
        Method & PSNR$\uparrow$ & SSIM$\uparrow$ & LPIPS$\downarrow$ & PSNR$\uparrow$ & SSIM$\uparrow$ & LPIPS$\downarrow$ \\
    \midrule
        SV4D~\cite{sv4d} & 19.65 & 0.883 & 0.137 & - & - & -  \\
        \hdashline
        Sequential Arch. & 5.935 & 0.444 & 0.583 &  22.74 & \cellcolor{custom_green!35}{{0.743}} & \cellcolor{custom_green!35}{{0.147}}\\
        Parallel Arch. & \cellcolor{custom_green!35}{{21.40}} & \cellcolor{custom_green!35}{{0.892}} & \cellcolor{custom_green!35}{{0.124}} & \cellcolor{custom_green!35}{{22.92}} & 0.742 & 0.148 \\
        Fused View \& Time Attn. &  \cellcolor{custom_green!75}{\textbf{22.49}} & \cellcolor{custom_green!75}{\textbf{0.909}} & \cellcolor{custom_green!75}{\textbf{0.113}} & \cellcolor{custom_green!75}{\textbf{23.15}} & \cellcolor{custom_green!75}{\textbf{0.752}} & \cellcolor{custom_green!75}{\textbf{0.142}} \\
    \hline
    \end{tabular}
    \vspace{-5mm}
\end{table}

\vspace{-3mm}
\subsection{Feedforward Reconstruction}
\label{subsec:FFReconstruction}
\begin{comment}
\noindent\textbf{Preliminary: VGGT.}
VGGT~\cite{wang2025vggt} is a large neural network for 3D scene understanding. It processes multiple images using a DINO encoder to generate image tokens, which are combined with learnable camera tokens for each view. These tokens are concatenated and refined through Alternating-Attention layers in a transformer backbone. The refined camera tokens are passed to a dedicated head to predict camera parameters: rotation, translation, and focal length. The image tokens go through DPT-based heads to output depth maps, 3D point maps, and dense tracking features. All 3D predictions are made in the coordinate frame of the first camera. This canonical frame is defined by normalizing all 3D outputs relative to the first camera and scaling by the average distance of ground truth points to the origin.
\end{comment}
\vspace{-2mm}
Our 4D video generator synthesizes visual content, which is then passed to our Gaussian-based~\cite{kerbl3Dgaussians} feedforward reconstruction model. This model operates on RGB frames only, since camera parameters are not provided.
To recover geometry and camera parameters, we use a pretrained VGGT model~\cite{wang2025vggt}. VGGT is a transformer-based neural network for 3D scene understanding that processes multi-view images using DINO-encoded tokens and learnable camera tokens. It predicts camera parameters and dense 3D outputs such as depth maps and point clouds, all in a canonical frame aligned to the first camera. We unproject depth maps into 3D point clouds, which serve as Gaussian centroids. A DPT-based head, called the Gaussian head, is trained to estimate the remaining Gaussian parameters: opacities, scales, and rotations. Colors are derived from rays in RGB space and refined with residuals predicted by the Gaussian head. An overview of this process is shown in Figure~\ref{fig:feedforward_reconstruction}.

\noindent\textbf{Camera Token Replacement.}
To extend the model to dynamic scenes, applying it independently to each frame causes inconsistent camera predictions. To enforce temporal consistency, we replace the camera tokens of all views at each timestep with those from the first timestep, after the VGGT transformer blocks. This ensures that all frames share the same predicted camera parameters and improves consistency over time.

\noindent\textbf{Gaussian Head.}
The Gaussian head predicts opacities, rotations, and scales from refined image tokens. Centroids are derived from unprojected depth maps, with a predicted 3-dimensional pose-refinement which is added to the unprojected depths (following Splatt3r~\cite{smart2024splatt3r}). We train this module using a reconstruction loss composed of MSE and LPIPS: $\mathcal{L}_\text{recon} = \mathcal{L}_\text{MSE} + \lambda_\text{LPIPS} \mathcal{L}_\text{LPIPS}$, where the perceptual loss encourages photometric fidelity~\cite{zhang2018perceptual}.

% \begin{equation}
% \mathcal{L}_\text{reconstruction} = \mathcal{L}_\text{MSE} + \lambda_\text{LPIPS} \mathcal{L}_\text{LPIPS},
% \end{equation}

\noindent\textbf{Temporal Attention.}
The VGGT backbone uses Alternating-Attention layers to mix global and frame-wise information. After each frame and global attention layer, we add a temporal attention layer to connect tokens across timesteps. This helps the model share information across time in dynamic scenes. The temporal attention layer is zero-initialized so the model's initial predictions match the original VGGT outputs.

\noindent\textbf{Training.}
The first training stage uses both synthetic and real-world static datasets. We include RealEstate10K~\cite{realestate10k}, DL3DV~\cite{ling2024dl3dv}, MVImageNet~\cite{yu2023mvimgnet}, Kubric~\cite{kubric} (only single-timestep samples), and ACID~\cite{infinite_nature_2020}. Each iteration samples scenes and views, predicts Gaussian parameters, renders the scene, and computes the loss. The VGGT backbone stays frozen to reduce memory use and allow larger batch sizes. This setup simplifies training by letting VGGT handle geometry and color, while the Gaussian head learns only residual parameters. We then train on dynamic Kubric to tune the temporal attention layers, while continuing to finetune the Gaussian head. We also reuse static datasets at this stage by copying multi-view samples across timesteps to create static 4D datasets, which helps prevent forgetting. In both stages, we use $4$ source views. For dynamic training, we sample $4$ views at $4$ timesteps each. The rest of the training hyperparameters follow BTimer~\cite{liang2024btimer}. More details are provided in the supplementary.

% Static and dynamic training use batch sizes of $14$ and $1$, respectively, and learning rates of $0.0002$ and $0.00002$. We sample uniformly across datasets in both stages. Static training runs for $20K$ iterations, and dynamic training runs for $15K$ iterations. We use the same hyperparameters for temporal attention as for global attention. The same hyperparameters as the depth head are also used for the Gaussian head, except the output dimension is set to $14$: $3$ for position refinement, $1$ for opacity, $3$ for scales, $4$ for rotation (quaternion), and $3$ for color offsets. Color and pose offsets are added following Splatt3r~\cite{smart2024splatt3r}.

\vspace{-3mm}
\section{Experiments}
\vspace{-3mm}

\subsection{Synchronized Multi-View Video Generation Evaluation}
\vspace{-2mm}
\noindent\textbf{Evaluation datasets.}
We evaluate the 4D video generation capability across a combination of datasets:
1) \emph{Generated videos.}
We run Veo 2~\cite{veo2} to create 30 fixed-view videos, each prompted by a unique caption. To enforce a static viewpoint, we append the phrase \textit{“static shot. The camera is completely static and doesn't move at all.”} to each caption. We then extract the first frame of each generated video and duplicate it to create a static video. This static video is passed to ReCamMaster~\cite{recammaster} to produce a freeze-time video (reference freeze-time) of the scene. This process establishes the first column and first row of our 4D grid, which we keep consistent across all baselines that utilize it.
2) \textit{Objaverse.} Following SV4D, we collect 19 animated 3D assets from Objaverse~\cite{objaverse} that are not included in the training set.
3) \textit{Nvidia Dynamic Dataset.} This dataset~\cite{yoon2020novel} contains 9 dynamic scenes captured by 12 synchronized cameras, offering real-world multi-view data.

\noindent\textbf{Baselines.}
We compare against state-of-the-art video generation methods that either natively support synchronized multi-view generation or can be adapted to do so:
1) \emph{TrajectoryCrafter}~\cite{yu2025trajectorycrafterredirectingcameratrajectory} is a representative baseline for point cloud-conditioned methods. As a 2-view model, it generates fixed target views conditioned on the reference fixed-view video. We use it to produce 8 distinct views per scene. 
2) \emph{ReCamMaster}~\cite{recammaster} generates a video with modified camera trajectory,  conditioned on a reference video that shares the first frame with the output. Since it is not directly suitable for multi-view generation, we adapt it in two variants: \emph{ReCamMaster-V1}: We construct a pseudo-static reference video by repeating the first frame for the first half, followed by the original freeze-view video. The target camera trajectory moves during the first half and remains static in the second. We retain only the second half of the output, yielding an approximately fixed-view rendering from a new viewpoint.
\emph{ReCamMaster-V2}: We generate independent freeze-time videos for each timestep by conditioning on static reference videos, created by repeating the corresponding frames of the input fixed-view video.
3) \emph{SynCamMaster}~\cite{syncammaster} \& \emph{4Real-Video}~\cite{4real-video} are both multi-view models.  Since the released SynCamMaster code does not support frame conditioning, we evaluate it using text conditioning.
4) \emph{SV4D}~\cite{sv4d} is a multi-view generation model trained on object-centric data.

\begin{table}[t]
\caption{Quantitative comparison with baselines on the generated video dataset. }
    \label{tab:video_gen}
    % \vspace{-2mm}
    \centering
    \scriptsize
    % \begin{adjustbox}{max width=\textwidth}
    \begin{tabular}{lc|ccccc}
    \toprule
             & Cross-View & \multicolumn{5}{c}{Cross-Time (VBench~\cite{vbench++})} \\
    Method   &  Met3R$\downarrow$~\cite{met3r}  & Flickering$\uparrow$ & Motion$\uparrow$ & Subject$\uparrow$ & Background $\uparrow$& Image$\uparrow$\\
    \midrule
TrajectoryCrafter~\cite{yu2025trajectorycrafterredirectingcameratrajectory} & 0.324 & 97.1 & 98.5 & 95.3 & 96.8 & \cellcolor{custom_green!75}{\textbf{67.6}}\\
    SynCamMaster~\cite{syncammaster} & 0.530 & \cellcolor{custom_green!75}{\textbf{99.3}} & \cellcolor{custom_green!75}{\textbf{99.5}} & 97.2 & 96.6 & 65.7 \\
    ReCamMaster-V1~\cite{recammaster} & 0.530 & 98.6 & \cellcolor{custom_green!35}{99.4} & 96.6 & 96.1 & 66.0\\
    ReCamMaster-V2~\cite{recammaster} & 0.194 & 94.7 & 91.2 & 90.7 & 93.6 & 65.4 \\
    4Real-Video~\cite{4real-video} & \cellcolor{custom_green!35}{0.192} & 98.7 & 99.2 & 94.4 & 96.5 & 64.4\\
    \hdashline
    w/o 2D Trans. Videos & 0.196 & \cellcolor{custom_green!75}{\textbf{99.3}} & \cellcolor{custom_green!75}{\textbf{99.5}} & \cellcolor{custom_green!75}{\textbf{98.0}} & \cellcolor{custom_green!75}{\textbf{98.7}} & 63.9 \\
    Full Method & \cellcolor{custom_green!75}{\textbf{0.173}} & \cellcolor{custom_green!35}{99.1} & \cellcolor{custom_green!75}{\textbf{99.5}} & \cellcolor{custom_green!35}{97.7} & \cellcolor{custom_green!35}{98.4} &  \cellcolor{custom_green!35}{66.2} \\
    \hline
    \end{tabular}
    % \end{adjustbox}
    \vspace{-4mm}
\end{table}

\begin{figure}[t]
    \centering
\includegraphics[width=\linewidth]{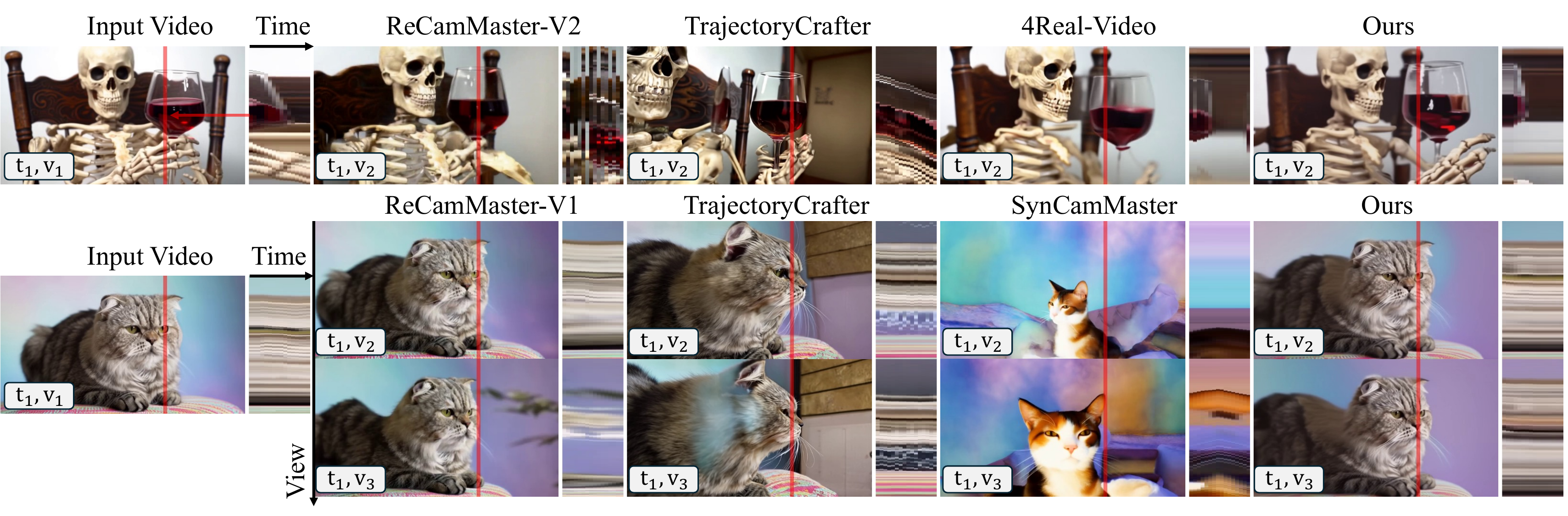}
\vspace{-6mm}
    \caption{Visual comparison of 4D video generation methods. Each image includes a temporal slice (right) along the red line to reveal temporal flickering. ReCamMaster-V2 shows strong flickering; ReCamMaster-V1 has inconsistent backgrounds across views. TrajectoryCrafter exhibits artifacts from noisy point clouds. 4Real-Video misses thin structures, and SynCamMaster produces inconsistent synthetic-style results. Our method achieves the best visual quality and consistency. }
    \label{fig:video_gen}
    \vspace{-2mm}
\end{figure}

\noindent\textbf{Evaluation protocal}. For all datasets, we prepared fixed-view videos as reference inputs and freeze-time videos to condition the view points for generation. For methods that do not support frame-level conditioning, we instead provide camera poses. On datasets with ground truth frames (Objaverse and NVIDIA Dynamic), we evaluate reconstruction quality using standard metrics: PSNR, SSIM, and LPIPS. For the generated video dataset, we subsample outputs into fixed-view and freeze-time videos. We assess multi-view consistency using Met3R~\cite{met3r}, and evaluate visual quality of the fixed-view videos using the widely adopted VBench~\cite{vbench++} metrics.

\noindent\textbf{Comparing model architectures.} We compare three model architecture variants (see Fig.\ref{fig:prelim_architecture}): \emph{sequential}, \emph{parallel}, and our proposed fused view-time attention. All models are trained under identical settings for 4,000 iterations with a batch size of 96. We evaluate their performance quantitatively on the Objaverse and NVIDIA Dynamic datasets (see Tab.\ref{tab:objaverse_and_nvidia}). Our proposed fused view-time attention consistently outperforms the other variants. The parallel architecture shows a noticeable drop in performance, suggesting that the introduction of bottleneck (synchronization) layers may limit model capacity. The sequential architecture learns significantly slower and fails to generate proper white backgrounds for Objaverse scenes (See the supplement for visualizations).

\noindent\textbf{Importance of 2D Transformed Videos.} As shown in Tab.~\ref{tab:video_gen}, removing 2D transformed videos from the training set leads to a noticeable drop in performance. This highlights the value of using abundant 2D video data for augmentation, especially when synthetic 4D data is limited.

\noindent\textbf{Comparing baselines.} On Objaverse, our method produces noticeably higher-quality results compared to SV4D, as showin in Tab.~\ref{tab:objaverse_and_nvidia} (see the supplement for visuals). SV4D often generates blurry frames. We attribute this to limitations of its base model and exclusive training on synthetic data.

On the generated dataset, our method consistently outperforms all baselines in both video quality and multi-view consistency (see Tab.\ref{tab:video_gen} and Fig.\ref{fig:video_gen}). The publicly released SynCamMaster model shows noticeable inconsistencies across views and exhibits a bias toward synthetic-style outputs. ReCamMaster-V1 struggles to maintain a static camera trajectory, and because each fixed-view video is generated independently, it lacks multi-view consistency. ReCamMaster-V2 achieves better multi-view alignment but suffers from temporal flickering. TrajectoryCrafter produces consistent outputs overall, but artifacts often emerge due to outliers in the conditioned point clouds, reducing visual fidelity. Lastly, 4Real-Video is constrained by its low-resolution, pixel-based model, results in degraded visual quality and frequent failure to render fine details, such as the fingers of the skeleton.

\begin{figure}[t]
    \centering    \includegraphics[width=1.0\linewidth]{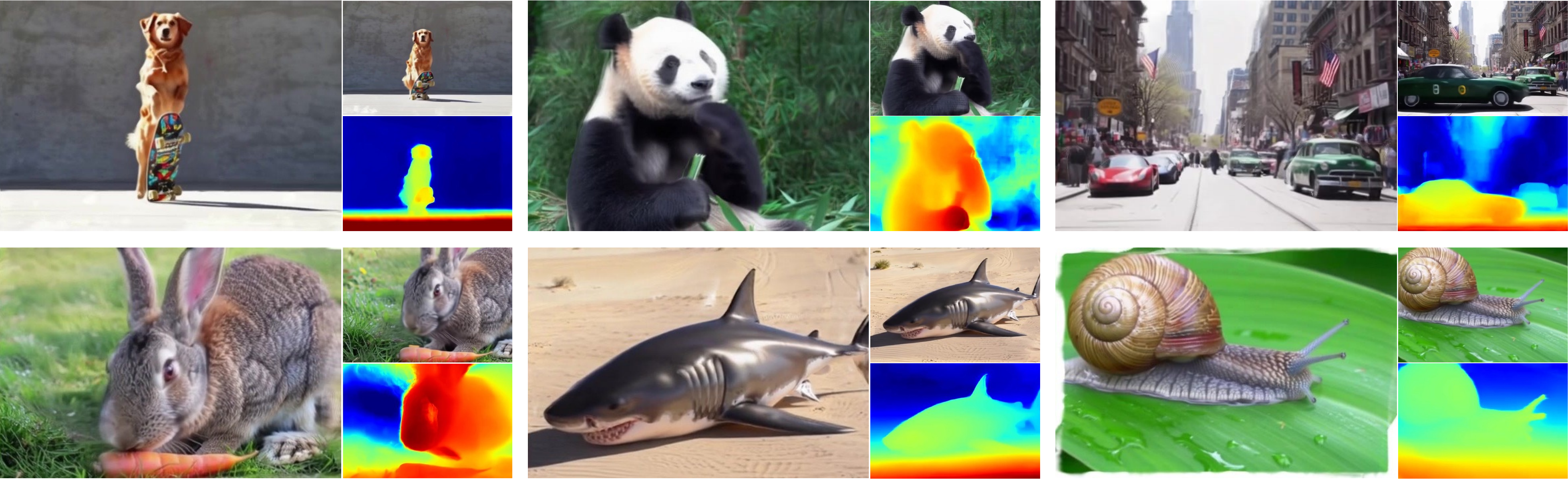}
    \vspace{-5mm}
    \caption{Color and depth renderings of Gaussians produced by inputting our generated 4D video grids into the reconstruction model.}
    \label{fig:gaussians_renderings_qualitative}
    \vspace{-4mm}
\end{figure}

\vspace{-2mm}
\subsection{Feedforward Reconstruction Evaluation}
\vspace{-2mm}
We evaluate our feedforward reconstruction model on its own, comparing it to state-of-the-art methods for static and dynamic scenes. See the supplement for details on the baselines, datasets, and metrics.

% \noindent\textbf{NVS for static scenes.} We first compare our method against the baselines on the task of static scene NVS. Figure~\ref{fig:gslrm_btimer_qualitative_llff} shows visual comparisons with GSLRM~\cite{gslrm2024} and BTimer~\cite{liang2024btimer} on the LLFF~\cite{mildenhall2019llff} dataset. The baselines use ground-truth camera calibration, while our method only takes RGB images as input. Despite this harder setup, our method produces higher quality renderings. It also avoids layering artifacts due to better underlying geometry. In contrast, BTimer struggles in this regard. When BTimer is finetuned and tested with 16 input images, it fails to align the per-pixel Gaussian predictions across views. This leads to visible layering artifacts, such as on the leaves of the fern. The static reconstruction model from BTimer uses the same architecture as GSLRM but is trained on more datasets. As a result, it generalizes better to unseen datasets like and LLFF, similar to our method. Please see the supplemental for additional visualizations, including on the Tanks \& Temples~\cite{Knapitsch2017tandt} dataset.

\noindent\textbf{NVS for static scenes.} 
Table~\ref{tab:gslrm_btimer_quantitative} shows quantitative results comparing our method with the baselines (See the supplement for visualizations). GSLRM and BTimer lack scene-scale standardization, so their performance can vary if camera parameters are manually scaled for each test scene (using a tedious grid search). To ensure fairness, we report results for GSLRM/BTimer with and without per-scene manual scale tuning (see the ``Manual Scale" column).
Even with tuning, these methods do not match our performance. Our method does not use manual tuning, nor does it rely on input/output camera parameters. Instead, it predicts all camera parameters. This adds difficulty, as it can affect both Gaussian predictions and camera accuracy. We use VGGT~\cite{wang2025vggt} to predict the output camera parameters by taking the first input frame and the target frames. These predictions are made in the coordinate system of the first input, which is also the frame of reference for our Gaussians.
Despite these challenges, our method still outperforms the baselines, even when they are manually tuned.

\begin{table}[t]
\caption{Quantitative comparison with baselines on static scene novel view synthesis.}
\label{tab:gslrm_btimer_quantitative}
% \vspace{-2mm}
\begin{adjustbox}{max width=\textwidth}
\centering
\scriptsize
\begin{tabular}{lccccccccc}
\toprule
\multicolumn{1}{c}{} &  &   &  & \multicolumn{3}{c}{Tanks \& Temples~\cite{Knapitsch2017tandt}} & \multicolumn{3}{c}{LLFF~\cite{mildenhall2019llff}} \\
Method & \# Inputs & Input Cams. & Manual Scale & PSNR$\uparrow$ & LPIPS$\downarrow$ & SSIM$\uparrow$ & PSNR$\uparrow$ & LPIPS$\downarrow$ & SSIM$\uparrow$ \\ \midrule
GSLRM~\cite{gslrm2024} & 4 & \textcolor{custom_red}{Yes} & \textcolor{custom_red}{Yes} & 15.78 & 0.3896 & 0.3385 & 14.42 & 0.4465 & 0.2980 \\
GSLRM~\cite{gslrm2024} & 16 & \textcolor{custom_red}{Yes} & \textcolor{custom_red}{Yes} & 16.21 & 0.4236 & 0.3528 & 14.85 & 0.4919 & 0.3222 \\
BTimer~\cite{liang2024btimer} (Static) & 4 & \textcolor{custom_red}{Yes} & \textcolor{custom_red}{Yes} & \cellcolor{custom_green!35}{20.62} & \cellcolor{custom_green!35}{0.1498} & 0.5762 & 16.40 & 0.2789 & 0.3669 \\
BTimer~\cite{liang2024btimer} (Static) & 16 & \textcolor{custom_red}{Yes} & \textcolor{custom_red}{Yes} & 20.45 & 0.1633 & 0.\cellcolor{custom_green!35}{5771} & \cellcolor{custom_green!35}{16.60} & 0.3225 & \cellcolor{custom_green!35}{0.3971} \\ \hdashline
GSLRM~\cite{gslrm2024} & 4 & \textcolor{custom_red}{Yes} & \textcolor{custom_green}{No} & 12.41 & 0.5933 & 0.3038 & 10.69 & 0.6182 & 0.2785 \\
GSLRM~\cite{gslrm2024} & 16 & \textcolor{custom_red}{Yes} & \textcolor{custom_green}{No} & 12.40 & 0.6138 & 0.3244 & 12.15 & 0.6323 & 0.2946 \\
BTimer~\cite{liang2024btimer} (Static) & 4 & \textcolor{custom_red}{Yes} & \textcolor{custom_green}{No} & 16.48 & 0.2781 & 0.3933 & 14.24 & 0.3958 & 0.2830 \\
BTimer~\cite{liang2024btimer} (Static) & 16 & \textcolor{custom_red}{Yes} & \textcolor{custom_green}{No} & 17.13 & 0.2883 & 0.4477 & 14.63 & 0.4231 & 0.3142 \\
Splatt3r~\cite{smart2024splatt3r} & 2 & \textcolor{custom_green}{No} & \textcolor{custom_green}{No} & 12.50 & 0.4547 & 0.3363 & 12.64 & 0.4599 & 0.3055 \\
Ours & 4 & \textcolor{custom_green}{No} &  \textcolor{custom_green}{No} & 18.52 & 0.1699 & 0.5178 & 15.12 & \cellcolor{custom_green!35}{0.2778} & 0.3024 \\
Ours & 16 & \textcolor{custom_green}{No} &  \textcolor{custom_green}{No} & \cellcolor{custom_green!75}{\textbf{20.85}} & \cellcolor{custom_green!75}{\textbf{0.1464}} & \cellcolor{custom_green!75}{\textbf{0.6057}} & \cellcolor{custom_green!75}{\textbf{18.95}} & \cellcolor{custom_green!75}{\textbf{0.1919}} & \cellcolor{custom_green!75}{\textbf{0.4573}} \\ \bottomrule
\end{tabular}
\end{adjustbox}
\vspace{-2mm}
\end{table}

\begin{table}[h]
\caption{Comparison with baselines for dynamic NVS on the Neural3DVideo~\cite{li2022neural3dvideosynthesis} dataset.}
% \vspace{-2mm}
\label{tab:gslrm_btimer_dynamic_quantitative}
\centering
\scriptsize
\begin{tabular}{lcccccc}
\toprule
Method & Input Cams. & Manual Scale & PSNR$\uparrow$ & LPIPS$\downarrow$ & SSIM$\uparrow$ \\ \midrule
GSLRM~\cite{gslrm2024} & \textcolor{custom_red}{Yes} & \textcolor{custom_red}{Yes} & 20.54 & 0.1934 & 0.6346 \\
BTimer~\cite{liang2024btimer} (monocular) & \textcolor{custom_red}{Yes} & \textcolor{custom_red}{Yes} & 9.65 & 0.4310 & 0.3907 \\
BTimer~\cite{liang2024btimer} (multi-view) & \textcolor{custom_red}{Yes} & \textcolor{custom_red}{Yes} & \cellcolor{custom_green!35}{21.55} & \cellcolor{custom_green!35}{0.1213} & \cellcolor{custom_green!75}{\textbf{0.6412}} \\ \hdashline
GSLRM~\cite{gslrm2024} & \textcolor{custom_red}{Yes} & \textcolor{custom_green}{No} & 12.31 & 0.5866 & 0.3553 \\
BTimer~\cite{liang2024btimer} (monocular) & \textcolor{custom_red}{Yes} & \textcolor{custom_green}{No} & 9.40 & 0.5898 & 0.3006 \\
BTimer~\cite{liang2024btimer} (multi-view) & \textcolor{custom_red}{Yes} & \textcolor{custom_green}{No} & 19.56 & 0.1693 & 0.6312 \\
Ours & \textcolor{custom_green}{No} & \textcolor{custom_green}{No} & \cellcolor{custom_green!75}{\textbf{21.63}} & \cellcolor{custom_green!75}{\textbf{0.1200}} & \cellcolor{custom_green!35}{0.6375} \\ \bottomrule
\end{tabular}
\vspace{-3mm}
\end{table}

\begin{figure}[h!]
    \centering
    \includegraphics[width=1.0\linewidth]{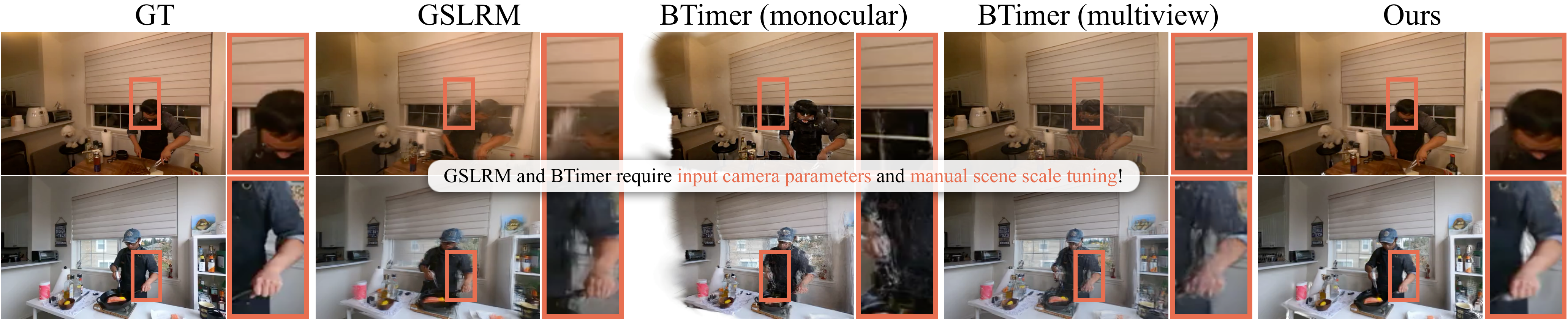}
    \vspace{-6mm}
    \caption{Qualitative comparison of our feedforward reconstruction model with the baselines on novel view renderings of dynamic scenes from the Neural3DVideo~\cite{li2022neural3dvideosynthesis} dataset.}
    \label{fig:gslrm_btimer_qualitative_dynamic}
    \vspace{-3mm}
\end{figure}

\begin{wraptable}{r}{0.36\linewidth} % 'r' for right, and width set to 7cm
    \centering
    \scriptsize
    \vspace{-6mm}
    \caption{Ablation of our feedforward reconstruction model on the test set of the dynamic Kubric~\cite{kubric} dataset.}
    \label{tab:reconstruction_ablation_kubric}
    \begin{tabular}{ll}
        \toprule
        Method                            & PSNR  \\
        \midrule
        w/o cam. token replacement        & 22.48 \\
        w/o temporal attention            & \cellcolor{custom_green!35}{22.60} \\
        Full model                        & \cellcolor{custom_green!75}{\textbf{23.39}} \\
        \bottomrule
    \end{tabular}
    \vspace{-3mm}
\end{wraptable}

\noindent\textbf{NVS for dynamic scenes.} We compare our model to baselines on novel view synthesis for dynamic scenes using the Neural3DVideo~\cite{li2022neural3dvideosynthesis} dataset, which none of the baselines were trained on. Figure~\ref{fig:gslrm_btimer_qualitative_dynamic} shows a visual comparison. GSLRM is designed for static scenes and is applied here by processing each timestep separately.
BTimer~\cite{liang2024btimer} claims to support monocular video input but can also take a 4D grid. For fairness, we evaluate both versions: monocular and multi-view. The monocular version cannot generate views outside the input trajectory and struggles with large camera motions.
As in static scenes, our method faces a harder task. It predicts both the Gaussians and the camera parameters for input/output views. Still, it produces sharper, more accurate results. Table~\ref{tab:gslrm_btimer_dynamic_quantitative} confirms this with better metrics, even when the baselines are manually tuned per scene. In this experiment, for each scene, we use 4 views as input and the rest as target. 
We also perform an ablation study by removing the camera token replacement and the temporal attention components, as shown in Table~\ref{tab:reconstruction_ablation_kubric}. The results indicate that both components contribute to improved quality in the rendered novel views.

\vspace{-2mm}
\section{Conclusion}
\vspace{-3mm}
We presented a two-stage framework for 4D video generation that leverages large-scale 4D video diffusion models and a feedforward reconstruction network to produce dynamic Gaussian splats from synchronized multi-view videos. While our method achieves state-of-the-art performance across multiple benchmarks, several \textbf{limitations} remain. First, the current design does not support full 360-degree scene generation. Second, although multi-view consistency is improved over prior methods, some layering artifacts can still appear in the reconstructed Gaussian splats. Third, inference remains computationally expensive, requiring approximately 4 minutes to generate 8 views and 29 frames on a single A100 GPU. Future work may explore model distillation to improve inference speed and expand scene coverage.
%Our proposed fused view-time attention introduces no additional parameters to the base model and enables efficient, consistent multi-view and temporal generation. The reconstruction stage further improves efficiency by predicting both camera poses and dynamic representations in a single forward pass.

\setcounter{page}{1}   
\appendix

% \clearpage
% \newpage

% \input{checklist}

% \clearpage
% \newpage

\setcounter{figure}{0}   % start figures at 1 again
\setcounter{table}{0}    % start tables  at 1 again
\renewcommand\thefigure{S\arabic{figure}} % Fig. S1, S2, …
\renewcommand\thetable {S\arabic{table}}  % Table S1, S2, …

% \section*{Appendix: Fused View-Time Attention and Feedforward Reconstruction for 4D Scene Generation}

\begin{center}
{\LARGE\bfseries
Appendix
}\par
\end{center}
% \vspace{1em}

\section{Visual results for comparing architectures for 4D video generation}

\begin{figure}[b]
    \centering
    \includegraphics[width=\linewidth]{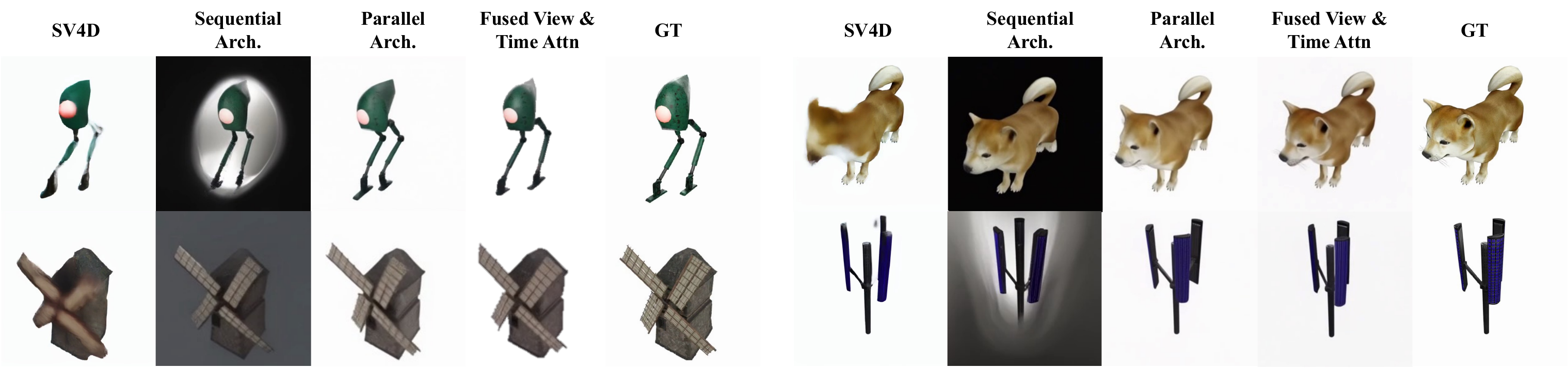}
    \caption{Qualitative comparison of the outputs of our proposed architecture (fused view-time attention) with our implementation of sequential and parallel architectures and SV4D~\cite{sv4d}. }
    \label{fig:objaverse}
\end{figure}

Figure~\ref{fig:objaverse} presents a qualitative comparison of outputs from various model architectures: \emph{SV4D}~\cite{sv4d}, \emph{sequential}, \emph{parallel}, and our proposed \emph{fused view-time attention} architecture. Each example showcases a frame from a 4D video. The \emph{fused view-time attention} model produces the most consistent and realistic results, closely resembling the ground truth in both shape and appearance. In contrast, the \emph{sequential} architecture exhibits lighting artifacts and fails to maintain a clean background, particularly in the Objaverse scenes. The \emph{parallel} architecture performs better but still shows noticeable temporal instability and degradation in fine details. \emph{SV4D} suffers from significant blurriness and structural distortions, underscoring the advantages of joint view-time modeling in our proposed approach. Please refer to Table~\ref{tab:objaverse_and_nvidia} from the main paper for a quantitative comparison. The results for the sequential and parallel architectures stem from our own reimplementation of these architectures, so that all architectures use the same video model as backbone for a fair comparison (besides SV4D).

\begin{table}[b]
\caption{Runtime (seconds) and peak GPU memory (GBs) required by our feed-forward reconstruction network during inference on static and dynamic scene sequences, reported for varying numbers of input camera views and timesteps. \textit{OOM} means the model has ran out of memory.}
\label{tab:feedforward_timings_and_memory}
\centering
\scriptsize
\begin{tabular}{lcccccccc}
\toprule
               & \multicolumn{2}{c}{1 Timestep (Static)}                         & \multicolumn{2}{c}{4 Timesteps} & \multicolumn{2}{c}{8 Timesteps} & \multicolumn{2}{c}{16 Timesteps} \\
\# Input Views & \multicolumn{1}{c}{Time (s)} & \multicolumn{1}{c}{Mem. (GB)} & Time (s)     & Mem. (GB)     & Time (s)     & Mem. (GB)     & Time (s)     & Mem. (GB) \\ \midrule
2  & 0.1779 & 7.204  & 0.4313 & 10.282 & 0.6850 & 13.709 & 1.2317 & 20.571 \\
4  & 0.2044 & 7.885  & 0.6467 & 13.192 & 1.1712 & 18.528 & 2.1725 & 32.213 \\
8  & 0.2742 & 9.319  & 1.3009 & 18.933 & 2.4204 & 31.011 & 5.6977 & 60.172 \\
16 & 0.5566 & 10.817 & 2.7396 & 24.989 & 5.2527 & 43.123 & \textit{OOM} & \textit{OOM} \\ \bottomrule
\end{tabular}
\end{table}

\section{Additional details on 4D video diffusion model}
\noindent\textbf{Architecture details.} The base video model is a latent diffusion model built on a DiT backbone, consisting of 32 DiT blocks with a hidden size of 4096, and a total of 11B learnable parameters. We use rotary positional embedding (RoPE) for its relative encoding properties and strong generalization across varying resolutions and durations. The model employs a convolutional autoencoder similar to that in CogVideo~\cite{cogvideox}, achieving 8$\times$ compression in the spatial dimensions and 4$\times$ in the temporal dimension. We fine-tune our 4D model using videos at a resolution of 144×256, and observe that it generalizes well to higher resolutions (e.g., 288×512) and longer durations without additional training. 

\noindent\textbf{Training data composition.} Training Data Composition. Our training set comprises a combination of synthetic 4D data from Objaverse and Kubric, 2D transformed videos, and videos of static scenes. Each training batch consists of 40\% Objaverse data, 20\% Kubric, 20\% 2D transformed videos, and 20\% static scene videos. For the static scenes, we duplicate and stack frames to construct a 4D video structure, although no actual object motion is present. To prevent the model from learning a trivial solution that simply replicates the first frame across all views, we find it necessary to remove frame conditioning when using freeze-time videos. Otherwise, the model tends to ignore viewpoint variation and fails to capture meaningful temporal dynamics. In addition, we observe that randomly reversing the order of viewpoints serves as an effective augmentation strategy that improves the model’s generalization capability.

\noindent\textbf{Training details.} Training Setup. We train the model on 48 A100 GPUs with a batch size of 96, using sequences of 8 views and 29 frames. The learning rate is set to $1e-4$
  with a warm-up schedule. The model converges quickly and begins producing plausible results after approximately 2000 iterations. We switch to fine-tuning the model on sequences with 8 views and 61 frames at 4000 iterations and the finetuning continues for additional 2000 iterations. We observe that the trained model generalizes well to sequences with varying numbers of frames, even when they differ from the configuration used during training.

\noindent\textbf{Sampling Strategy and Classifier-Free Guidance.} We adopt a rectified flow sampler consistent with our base video model. In the setting where both freeze-time and fixed-view videos are provided as input, we find that classifier-free guidance (CFG) is unnecessary, as it does not yield noticeable improvements in output quality. Under this configuration, our model is capable of generating high-quality results with a small number of diffusion steps—for example, as shown in Tab.~\ref{tab:score_per_step}, using only 4 steps already produces temporally consistent outputs, particularly in background regions. Further refinement of the foreground, especially in areas with larger motion, occurs with additional steps. This suggests that our model could potentially benefit from distillation techniques aimed at reducing the number of inference steps.

However, when only a single video is used as input, CFG remains essential. In this case, the model relies more heavily on the input text to resolve ambiguities during generation.

\noindent\textbf{Other variants of positional embedding for 4D video}  In addition to the design proposed in the main paper, we explored alternative formulations for converting 4D coordinates into 3D positional embeddings. Notably, we experimented with the transformation $(v,t,x,y)\rightarrow (v+t,x,y)$, based on the intuition that temporal indices are consecutive across rows and columns of the frame matrix. This mapping preserves the structural assumptions of the pretrained base video model.

Empirically, this variant performs comparably to our proposed embedding scheme when both freeze-time and fixed-view videos are used as input. However, when only one of the two input types is provided, the results become less stable. We attribute this to the ambiguity introduced by the $(v+t,x,y)$
 formulation, which leads to duplicated or symmetric positions in the frame grid. Specifically, positions become indistinguishable along the diagonal of the view-time plane, making it difficult for the model to differentiate between the temporal and view dimensions. As a result, the model must rely more heavily on the input frames themselves to infer the underlying structure.

\begin{table}[]
    \centering
    \begin{tabular}{cc|c|ccccc}
    \toprule
             &  & Cross-View & \multicolumn{5}{c}{Cross-Time (VBench~\cite{vbench++})} \\
    \# Step & Time (s)  &  Met3R$\downarrow$~\cite{met3r}  & Flickering$\uparrow$ & Motion$\uparrow$ & Subject$\uparrow$ & Background $\uparrow$& Image$\uparrow$\\
    \midrule
    4  & 47.2 & 0.187 & 94.6 & 97.8 & 96.3 & 97.7 & 64.7 \\
    8 & 89.4 & 0.184 & 94.5 & 97.7 & 96.5 & 97.7 & 65.6 \\
    16 & 173.8 & 0.183 & 94.4 & 97.7 & 96.6 & 97.7 & 65.7 \\
    40 & 472.0 & 0.173 & 99.1 & 99.5 & 97.7 & 98.4 &  66.2\\
    \bottomrule
    \end{tabular}
    \caption{Cross-view consistency and Cross-time quality assement for generation with different diffusion steps. Runtime is estimated for generating 4D videos with 8 views and 61 timestamps, in total 488 frames.}
    \label{tab:score_per_step}
\end{table}

\section{Additional details on the feedforward reconstruction model}

Static and dynamic training use batch sizes of $14$ and $1$, respectively, and learning rates of $0.0002$ and $0.00002$. We sample uniformly across datasets in both stages. Static training runs for $20K$ iterations, and dynamic training runs for $15K$ iterations. We use the same hyperparameters for temporal attention as for global attention in VGGT~\cite{wang2025vggt}. The same hyperparameters as VGGT's depth head are also used for the Gaussian head, except the output dimension is set to $14$: $3$ for position refinement, $1$ for opacity, $3$ for scales, $4$ for rotation (quaternion), and $3$ for color offsets. Color and pose offsets are added following Splatt3r~\cite{smart2024splatt3r}.

Table~\ref{tab:feedforward_timings_and_memory} provides an overview of the time and GPU memory usage required to run our feedforward reconstruction model on both dynamic and static datasets. Our model is capable of producing Gaussians for static and dynamic scenes within seconds. These metrics are calculated on an Nvidia A100 GPU. This experiment is conducted using inputs with a resolution of $350 \times 518$, following the standard input dimensions of VGGT.

% \begin{figure}[t]
%     \centering
%     \includegraphics[width=0.9\linewidth]{figures/gslrm_btimer_qualitative_llff_short_v2.pdf}
%     \caption{Qualitative comparison of our feedforward reconstruction model with the baselines on novel view renderings of static scenes from the LLFF~\cite{mildenhall2019llff} dataset.}
%     \label{fig:gslrm_btimer_qualitative_llff}
% \end{figure}

% \begin{figure}
%     \centering
%     \includegraphics[width=1.0\linewidth]{figures/gslrm_btimer_qualitative.pdf}
%     \caption{...}
%     \label{fig:gslrm_btimer_qualitative}
% \end{figure}

\section{Visual results for static scene novel view synthesis}

\begin{figure}
    \centering
    \includegraphics[width=1.0\linewidth]{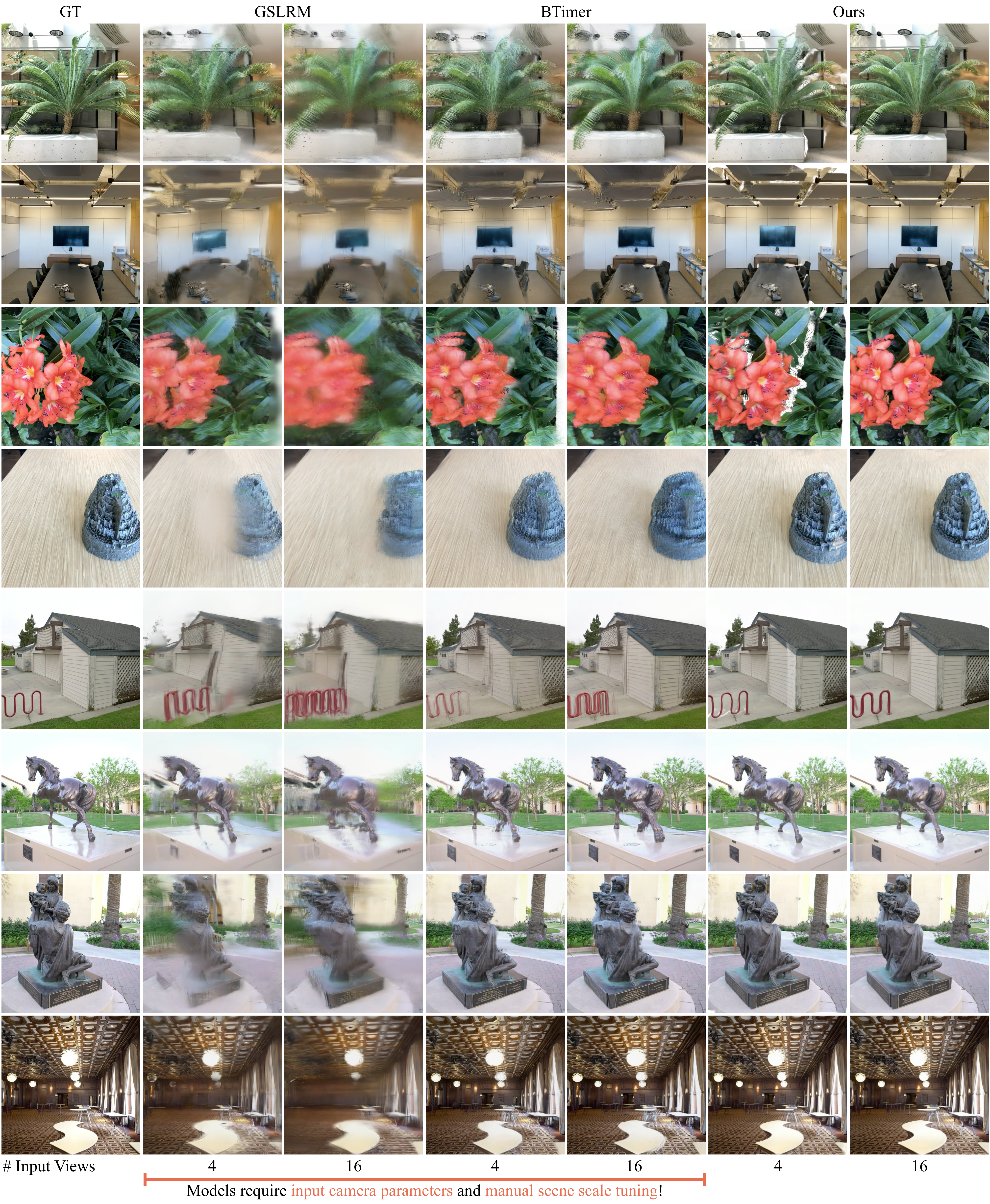}
    \caption{Qualitative comparison of our renderings with the baselines GSLRM~\cite{gslrm2024} and BTimer~\cite{liang2024btimer} on the task of novel view synthesis for static scenes. Each method includes two variations, using 4 and 16 input views. Note that all variations of GSLRM and BTimer require \textcolor{custom_red}{input camera parameters} and \textcolor{custom_red}{manual scene scale tuning}. }
    \label{fig:gslrm_btimer_qualitative_full}
\end{figure}

\begin{figure}
    \centering
    \includegraphics[width=1.0\linewidth]{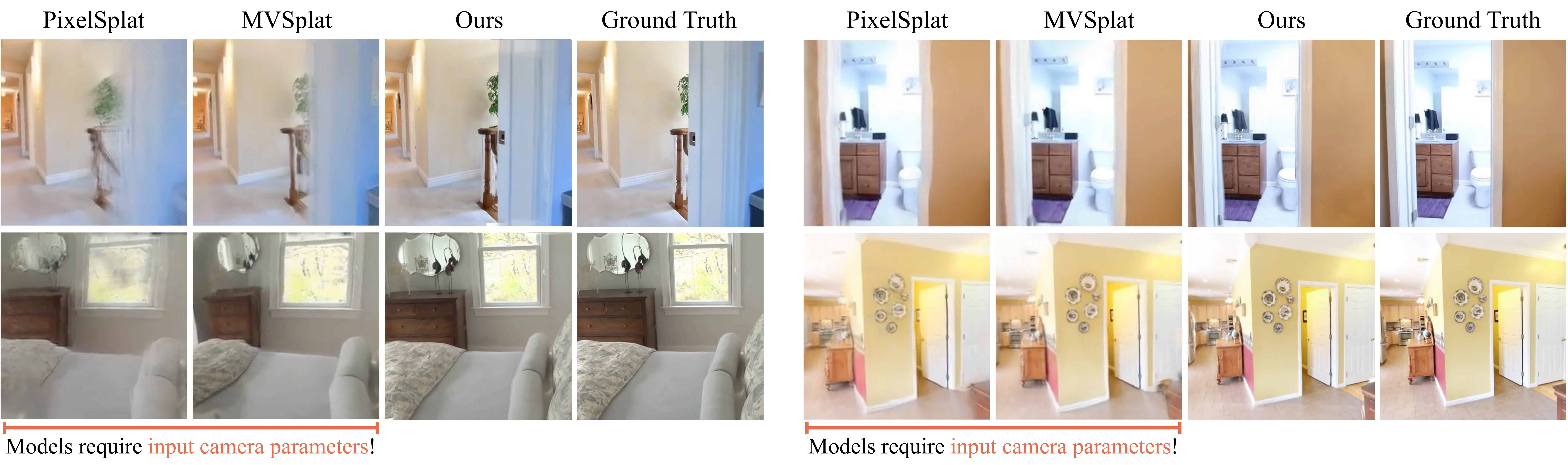}
    \vspace{-6mm}
    \caption{Qualitative comparison of our renderings with the baselines PixelSplat~\cite{charatan23pixelsplat} and MVSPlat~\cite{chen2024mvsplat} on the task of novel view synthesis for static scenes. Note that the baselines require \textcolor{custom_red}{input camera parameters}, whereas our method infers the camera parameters from the input images.}
    \label{fig:pixelsplat_mvsplat_qualitative}
\end{figure}

Figure~\ref{fig:gslrm_btimer_qualitative_full} supports the quantitative results in Table~\ref{tab:gslrm_btimer_quantitative} from the main paper. We compare our method with GSLRM~\cite{gslrm2024} and
BTimer~\cite{liang2024btimer} on LLFF~\cite{mildenhall2019llff} and Tanks \& Temples~\cite{Knapitsch2017tandt} scenes. The baselines need ground-truth camera poses and a per-scene scale search, while our method predicts all camera parameters and uses no manual tuning.  
GSLRM and BTimer are trained with a photometric loss only, so their per-view Gaussians do not stay aligned when the input set grows. With 16 input views the misalignment causes layering artifacts on fine details, such as the fern leaves, and on thin parts like the back leg of the \textit{Horse} statue.   Our model avoids these artifacts, matching the gains in PSNR, SSIM, and LPIPS
reported in the table.
In Fig~\ref{fig:pixelsplat_mvsplat_qualitative}, we also compare our method to PixelSplat~\cite{charatan23pixelsplat} and MVSplat~\cite{chen2024mvsplat} on the RealEstate10K~\cite{realestate10k} dataset. Our method produces visuals that more closely match the ground truth. Note that PixelSplat and MVSplat are trained specifically on RealEstate10K, so we compare on this dataset for fairness.

\begin{figure}[t]
    \centering
    \includegraphics[width=0.9\linewidth]{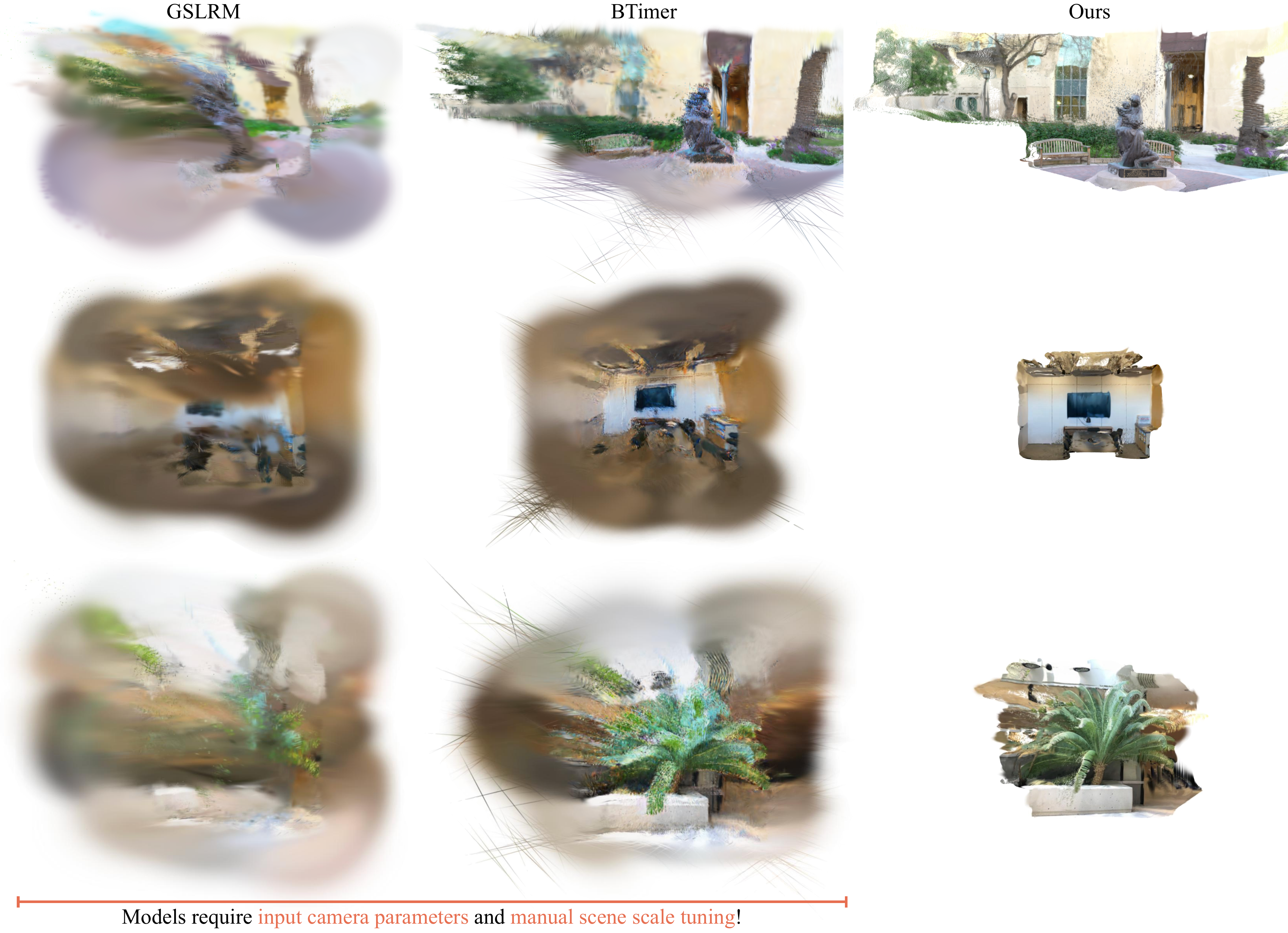}
    \caption{Qualitative comparison of our results with the baselines for novel view synthesis of static scenes, where the target camera deviates significantly  from the input trajectory.}
    \label{fig:extreme_camera_change_comparison}
\end{figure}

The difference between our renderings and those from the baselines is especially apparent when the target camera differs significantly from the input trajectory. Figure~\ref{fig:extreme_camera_change_comparison} shows renderings from our model compared to the baselines when the camera is moved backward and far from the set of input frames. Notably, our model is much better suited for view extrapolation, in part because it incorporates superior geometric priors, whereas the baselines rely solely on photometric losses.

\section{Dataset details}

Table~\ref{tab:datasets} provides an overview of the datasets used to train our models, summarizing key characteristics such as the presence of dynamic content, the type of content (object-centric or scene-level), the domain (real or synthetic), the approximate number of scenes, and the associated licenses. These datasets span a range of scenarios and content types, offering a diverse foundation for training models in our experiments.

\begin{table}[t]
\centering
\caption{Specification and licenses for the datasets used to train our models. }
\label{tab:datasets}
\scriptsize
\begin{tabular}{lcllll}
\toprule
\textbf{Dataset} & \textbf{Dynamic} & \textbf{Content} & \textbf{Domain} & \textbf{\# Scenes} & \textbf{License} \\ \midrule
RealEstate10K~\cite{realestate10k}  &   &  \textit{Scene} & Real & 80K & CC-BY (per video) \\
MVImageNet~\cite{yu2023mvimgnet}   &   &  \textit{Object}  & Real & 220K & Custom (password-protected) \\
DL3DV~\cite{ling2024dl3dv} &   &  \textit{Scene}  & Real & 10K & NonCommercial (custom terms) \\
Kubric~\cite{kubric}   &  \checkmark &   \textit{Object}+\textit{Scene}  & Synthetic & 3K & Apache 2.0 \\
Dynamic Objaverse~\cite{objaverse}  &  \checkmark &  \textit{Object}  & Synthetic &  & ODC-By v1.0 (mixed per object) \\
\bottomrule
\end{tabular}
\end{table}

\section{Broader impact}

By supporting 4D content creation, our method opens new possibilities in animation and visual effects. Nonetheless, careful consideration is required to prevent its exploitation for deceptive or harmful purposes, such as identity forgery.

\newpage

\bibliographystyle{unsrt}
\bibliography{neurips_2025.bbl}

\begin{thebibliography}{10}

\bibitem{4DiM}
Daniel Watson, Saurabh Saxena, Lala Li, Andrea Tagliasacchi, and David~J Fleet.
\newblock Controlling space and time with diffusion models.
\newblock {\em ICLR}, 2025.

\bibitem{cat4d}
Rundi Wu, Ruiqi Gao, Ben Poole, Alex Trevithick, Changxi Zheng, Jonathan~T. Barron, and Aleksander Holynski.
\newblock {CAT4D: Create Anything in 4D with Multi-View Video Diffusion Models}.
\newblock 2024.

\bibitem{genxd}
Yuyang Zhao, Chung-Ching Lin, Kevin Lin, Zhiwen Yan, Linjie Li, Zhengyuan Yang, Jianfeng Wang, Gim~Hee Lee, and Lijuan Wang.
\newblock Genxd: Generating any 3d and 4d scenes.
\newblock {\em ICLR}, 2025.

\bibitem{sun2024dimensionx}
Wenqiang Sun, Shuo Chen, Fangfu Liu, Zilong Chen, Yueqi Duan, Jun Zhang, and Yikai Wang.
\newblock Dimensionx: Create any 3d and 4d scenes from a single image with controllable video diffusion.
\newblock 2024.

\bibitem{gen3c}
Xuanchi Ren, Tianchang Shen, Jiahui Huang, Huan Ling, Yifan Lu, Merlin Nimier-David, Thomas MÃ¼ller, Alexander Keller, Sanja Fidler, and Jun Gao.
\newblock Gen3c: 3d-informed world-consistent video generation with precise camera control.
\newblock In {\em CVPR}, 2025.

\bibitem{yu2025trajectorycrafterredirectingcameratrajectory}
Mark YU, Wenbo Hu, Jinbo Xing, and Ying Shan.
\newblock Trajectorycrafter: Redirecting camera trajectory for monocular videos via diffusion models, 2025.

\bibitem{van2024generative}
Basile Van~Hoorick, Rundi Wu, Ege Ozguroglu, Kyle Sargent, Ruoshi Liu, Pavel Tokmakov, Achal Dave, Changxi Zheng, and Carl Vondrick.
\newblock Generative camera dolly: Extreme monocular dynamic novel view synthesis.
\newblock In {\em ECCV}, 2024.

\bibitem{recammaster}
Jianhong Bai, Menghan Xia, Xiao Fu, Xintao Wang, Lianrui Mu, Jinwen Cao, Zuozhu Liu, Haoji Hu, Xiang Bai, Pengfei Wan, and Di~Zhang.
\newblock Recammaster: Camera-controlled generative rendering from a single video, 2025.

\bibitem{cvd}
Zhengfei Kuang, Shengqu Cai, Hao He, Yinghao Xu, Hongsheng Li, Leonidas Guibas, and Gordon. Wetzstein.
\newblock Collaborative video diffusion: Consistent multi-video generation with camera control.
\newblock 2024.

\bibitem{human4dit}
Ruizhi Shao, Youxin Pang, Zerong Zheng, Jingxiang Sun, and Yebin Liu.
\newblock Human4dit: Free-view human video generation with 4d diffusion transformer.
\newblock 2024.

\bibitem{vividzoo}
Bing Li, Cheng Zheng, Wenxuan Zhu, Jinjie Mai, Biao Zhang, Peter Wonka, and Bernard Ghanem.
\newblock Vivid-zoo: Multi-view video generation with diffusion model, 2024.

\bibitem{4diffusion}
Haiyu Zhang, Xinyuan Chen, Yaohui Wang, Xihui Liu, Yunhong Wang, and Yu~Qiao.
\newblock 4diffusion: Multi-view video diffusion model for 4d generation.
\newblock 2024.

\bibitem{sv4d}
Yiming Xie, Chun-Han Yao, Vikram Voleti, Huaizu Jiang, and Varun Jampani.
\newblock Sv4d: Dynamic 3d content generation with multi-frame and multi-view consistency.
\newblock 2024.

\bibitem{syncammaster}
Jianhong Bai, Menghan Xia, Xintao Wang, Ziyang Yuan, Xiao Fu, Zuozhu Liu, Haoji Hu, Pengfei Wan, and Di~Zhang.
\newblock Syncammaster: Synchronizing multi-camera video generation from diverse viewpoints, 2024.

\bibitem{4real-video}
Chaoyang Wang, Peiye Zhuang, Tuan~Duc Ngo, Willi Menapace, Aliaksandr Siarohin, Michael Vasilkovsky, Ivan Skorokhodov, Sergey Tulyakov, Peter Wonka, and Hsin-Ying Lee.
\newblock 4real-video: Learning generalizable photo-realistic 4d video diffusion, 2024.

\bibitem{cogvideox}
Zhuoyi Yang, Jiayan Teng, Wendi Zheng, Ming Ding, Shiyu Huang, Jiazheng Xu, Yuanming Yang, Wenyi Hong, Xiaohan Zhang, Guanyu Feng, et~al.
\newblock Cogvideox: Text-to-video diffusion models with an expert transformer.
\newblock 2024.

\bibitem{wan2025}
Ang Wang, Baole Ai, Bin Wen, Chaojie Mao, Chen-Wei Xie, Di~Chen, Feiwu Yu, Haiming Zhao, Jianxiao Yang, Jianyuan Zeng, Jiayu Wang, Jingfeng Zhang, Jingren Zhou, Jinkai Wang, Jixuan Chen, Kai Zhu, Kang Zhao, Keyu Yan, Lianghua Huang, Mengyang Feng, Ningyi Zhang, Pandeng Li, Pingyu Wu, Ruihang Chu, Ruili Feng, Shiwei Zhang, Siyang Sun, Tao Fang, Tianxing Wang, Tianyi Gui, Tingyu Weng, Tong Shen, Wei Lin, Wei Wang, Wei Wang, Wenmeng Zhou, Wente Wang, Wenting Shen, Wenyuan Yu, Xianzhong Shi, Xiaoming Huang, Xin Xu, Yan Kou, Yangyu Lv, Yifei Li, Yijing Liu, Yiming Wang, Yingya Zhang, Yitong Huang, Yong Li, You Wu, Yu~Liu, Yulin Pan, Yun Zheng, Yuntao Hong, Yupeng Shi, Yutong Feng, Zeyinzi Jiang, Zhen Han, Zhi-Fan Wu, and Ziyu Liu.
\newblock Wan: Open and advanced large-scale video generative models.
\newblock 2025.

\bibitem{stepfun}
Haoyang Huang, Guoqing Ma, Nan Duan, Xing Chen, Changyi Wan, Ranchen Ming, Tianyu Wang, Bo~Wang, Zhiying Lu, Aojie Li, Xianfang Zeng, Xinhao Zhang, Gang Yu, Yuhe Yin, Qiling Wu, Wen Sun, Kang An, Xin Han, Deshan Sun, Wei Ji, Bizhu Huang, Brian Li, Chenfei Wu, Guanzhe Huang, Huixin Xiong, Jiaxin He, Jianchang Wu, Jianlong Yuan, Jie Wu, Jiashuai Liu, Junjing Guo, Kaijun Tan, Liangyu Chen, Qiaohui Chen, Ran Sun, Shanshan Yuan, Shengming Yin, Sitong Liu, Wei Chen, Yaqi Dai, Yuchu Luo, Zheng Ge, Zhisheng Guan, Xiaoniu Song, Yu~Zhou, Binxing Jiao, Jiansheng Chen, Jing Li, Shuchang Zhou, Xiangyu Zhang, Yi~Xiu, Yibo Zhu, Heung-Yeung Shum, and Daxin Jiang.
\newblock Step-video-ti2v technical report: A state-of-the-art text-driven image-to-video generation model, 2025.

\bibitem{moviegen}
Adam Polyak, Amit Zohar, Andrew Brown, Andros Tjandra, Animesh Sinha, Ann Lee, Apoorv Vyas, Bowen Shi, Chih-Yao Ma, Ching-Yao Chuang, David Yan, Dhruv Choudhary, Dingkang Wang, Geet Sethi, Guan Pang, Haoyu Ma, Ishan Misra, Ji~Hou, Jialiang Wang, Kiran Jagadeesh, Kunpeng Li, Luxin Zhang, Mannat Singh, Mary Williamson, Matt Le, Matthew Yu, Mitesh~Kumar Singh, Peizhao Zhang, Peter Vajda, Quentin Duval, Rohit Girdhar, Roshan Sumbaly, Sai~Saketh Rambhatla, Sam Tsai, Samaneh Azadi, Samyak Datta, Sanyuan Chen, Sean Bell, Sharadh Ramaswamy, Shelly Sheynin, Siddharth Bhattacharya, Simran Motwani, Tao Xu, Tianhe Li, Tingbo Hou, Wei-Ning Hsu, Xi~Yin, Xiaoliang Dai, Yaniv Taigman, Yaqiao Luo, Yen-Cheng Liu, Yi-Chiao Wu, Yue Zhao, Yuval Kirstain, Zecheng He, Zijian He, Albert Pumarola, Ali Thabet, Artsiom Sanakoyeu, Arun Mallya, Baishan Guo, Boris Araya, Breena Kerr, Carleigh Wood, Ce~Liu, Cen Peng, Dimitry Vengertsev, Edgar Schonfeld, Elliot Blanchard, Felix Juefei-Xu, Fraylie Nord, Jeff Liang, John Hoffman, Jonas
  Kohler, Kaolin Fire, Karthik Sivakumar, Lawrence Chen, Licheng Yu, Luya Gao, Markos Georgopoulos, Rashel Moritz, Sara~K. Sampson, Shikai Li, Simone Parmeggiani, Steve Fine, Tara Fowler, Vladan Petrovic, and Yuming Du.
\newblock Movie gen: A cast of media foundation models, 2024.

\bibitem{veo2}
Veo-Team:, Agrim Gupta, Ali Razavi, Andeep Toor, Ankush Gupta, Dumitru Erhan, Eleni Shaw, Eric Lau, Frank Belletti, Gabe Barth-Maron, Gregory Shaw, Hakan Erdogan, Hakim Sidahmed, Henna Nandwani, Hernan Moraldo, Hyunjik Kim, Irina Blok, Jeff Donahue, José Lezama, Kory Mathewson, Kurtis David, Matthieu~Kim Lorrain, Marc van Zee, Medhini Narasimhan, Miaosen Wang, Mohammad Babaeizadeh, Nelly Papalampidi, Nick Pezzotti, Nilpa Jha, Parker Barnes, Pieter-Jan Kindermans, Rachel Hornung, Ruben Villegas, Ryan Poplin, Salah Zaiem, Sander Dieleman, Sayna Ebrahimi, Scott Wisdom, Serena Zhang, Shlomi Fruchter, Signe Nørly, Weizhe Hua, Xinchen Yan, Yuqing Du, and Yutian Chen.
\newblock Veo 2.
\newblock 2024.

\bibitem{nerf}
Ben Mildenhall, Pratul~P Srinivasan, Matthew Tancik, Jonathan~T Barron, Ravi Ramamoorthi, and Ren Ng.
\newblock Nerf: Representing scenes as neural radiance fields for view synthesis.
\newblock {\em Communications of the ACM}, 2021.

\bibitem{kerbl3Dgaussians}
Bernhard Kerbl, Georgios Kopanas, Thomas Leimk{\"u}hler, and George Drettakis.
\newblock 3d gaussian splatting for real-time radiance field rendering.
\newblock {\em ToG}, 2023.

\bibitem{wang2025vggt}
Jianyuan Wang, Minghao Chen, Nikita Karaev, Andrea Vedaldi, Christian Rupprecht, and David Novotny.
\newblock Vggt: Visual geometry grounded transformer.
\newblock In {\em CVPR}, 2025.

\bibitem{dreamfusion}
Ben Poole, Ajay Jain, Jonathan~T Barron, and Ben Mildenhall.
\newblock Dreamfusion: Text-to-3d using 2d diffusion.
\newblock In {\em ICLR}, 2023.

\bibitem{prolificdreamer}
Zhengyi Wang, Cheng Lu, Yikai Wang, Fan Bao, Chongxuan Li, Hang Su, and Jun Zhu.
\newblock Prolificdreamer: High-fidelity and diverse text-to-3d generation with variational score distillation.
\newblock In {\em NeurIPS}, 2023.

\bibitem{sjc}
Haochen Wang, Xiaodan Du, Jiahao Li, Raymond~A Yeh, and Greg Shakhnarovich.
\newblock Score jacobian chaining: Lifting pretrained 2d diffusion models for 3d generation.
\newblock In {\em CVPR}, 2023.

\bibitem{fantasia3d}
Rui Chen, Yongwei Chen, Ningxin Jiao, and Kui Jia.
\newblock Fantasia3d: Disentangling geometry and appearance for high-quality text-to-3d content creation.
\newblock In {\em ICCV}, 2023.

\bibitem{magic3d}
Chen-Hsuan Lin, Jun Gao, Luming Tang, Towaki Takikawa, Xiaohui Zeng, Xun Huang, Karsten Kreis, Sanja Fidler, Ming-Yu Liu, and Tsung-Yi Lin.
\newblock Magic3d: High-resolution text-to-3d content creation.
\newblock In {\em CVPR}, 2023.

\bibitem{hifa}
Joseph Zhu and Peiye Zhuang.
\newblock Hifa: High-fidelity text-to-3d with advanced diffusion guidance.
\newblock In {\em ICLR}, 2023.

\bibitem{ldm}
Robin Rombach, Andreas Blattmann, Dominik Lorenz, Patrick Esser, and Bj{\"o}rn Ommer.
\newblock High-resolution image synthesis with latent diffusion models.
\newblock In {\em CVPR}, 2022.

\bibitem{imagen}
Chitwan Saharia, William Chan, Saurabh Saxena, Lala Li, Jay Whang, Emily~L Denton, Kamyar Ghasemipour, Raphael Gontijo~Lopes, Burcu Karagol~Ayan, Tim Salimans, et~al.
\newblock Photorealistic text-to-image diffusion models with deep language understanding.
\newblock {\em NeurIPS}, 2022.

\bibitem{zero123}
Ruoshi Liu, Rundi Wu, Basile Van~Hoorick, Pavel Tokmakov, Sergey Zakharov, and Carl Vondrick.
\newblock Zero-1-to-3: Zero-shot one image to 3d object.
\newblock In {\em ICCV}, 2023.

\bibitem{mvdream}
Yichun Shi, Peng Wang, Jianglong Ye, Mai Long, Kejie Li, and Xiao Yang.
\newblock Mvdream: Multi-view diffusion for 3d generation.
\newblock In {\em ICLR}, 2024.

\bibitem{4dfy}
Sherwin Bahmani, Ivan Skorokhodov, Victor Rong, Gordon Wetzstein, Leonidas Guibas, Peter Wonka, Sergey Tulyakov, Jeong~Joon Park, Andrea Tagliasacchi, and David~B Lindell.
\newblock 4d-fy: Text-to-4d generation using hybrid score distillation sampling.
\newblock In {\em CVPR}, 2024.

\bibitem{ayg}
Huan Ling, Seung~Wook Kim, Antonio Torralba, Sanja Fidler, and Karsten Kreis.
\newblock Align your gaussians: Text-to-4d with dynamic 3d gaussians and composed diffusion models.
\newblock In {\em CVPR}, 2024.

\bibitem{consistent4d}
Yanqin Jiang, Li~Zhang, Jin Gao, Weimin Hu, and Yao Yao.
\newblock Consistent4d: Consistent 360 $\{$$\backslash$deg$\}$ dynamic object generation from monocular video.
\newblock 2023.

\bibitem{dreamgaussian4d}
Jiawei Ren, Liang Pan, Jiaxiang Tang, Chi Zhang, Ang Cao, Gang Zeng, and Ziwei Liu.
\newblock Dreamgaussian4d: Generative 4d gaussian splatting.
\newblock 2023.

\bibitem{4dgen}
Yuyang Yin, Dejia Xu, Zhangyang Wang, Yao Zhao, and Yunchao Wei.
\newblock 4dgen: Grounded 4d content generation with spatial-temporal consistency.
\newblock 2023.

\bibitem{animate124}
Yuyang Zhao, Zhiwen Yan, Enze Xie, Lanqing Hong, Zhenguo Li, and Gim~Hee Lee.
\newblock Animate124: Animating one image to 4d dynamic scene.
\newblock 2023.

\bibitem{mav3d}
Uriel Singer, Shelly Sheynin, Adam Polyak, Oron Ashual, Iurii Makarov, Filippos Kokkinos, Naman Goyal, Andrea Vedaldi, Devi Parikh, Justin Johnson, et~al.
\newblock Text-to-4d dynamic scene generation.
\newblock 2023.

\bibitem{4real}
Heng Yu, Chaoyang Wang, Peiye Zhuang, Willi Menapace, Aliaksandr Siarohin, Junli Cao, Laszlo~A Jeni, Sergey Tulyakov, and Hsin-Ying Lee.
\newblock 4real: Towards photorealistic 4d scene generation via video diffusion models.
\newblock In {\em NeurIPS}, 2024.

\bibitem{animatediff}
Yuwei Guo, Ceyuan Yang, Anyi Rao, Zhengyang Liang, Yaohui Wang, Yu~Qiao, Maneesh Agrawala, Dahua Lin, and Bo~Dai.
\newblock Animatediff: Animate your personalized text-to-image diffusion models without specific tuning.
\newblock 2023.

\bibitem{imagenvideo}
Jonathan Ho, William Chan, Chitwan Saharia, Jay Whang, Ruiqi Gao, Alexey Gritsenko, Diederik~P Kingma, Ben Poole, Mohammad Norouzi, David~J Fleet, et~al.
\newblock Imagen video: High definition video generation with diffusion models.
\newblock 2022.

\bibitem{videocrafter}
Haoxin Chen, Menghan Xia, Yingqing He, Yong Zhang, Xiaodong Cun, Shaoshu Yang, Jinbo Xing, Yaofang Liu, Qifeng Chen, Xintao Wang, Chao Weng, and Ying Shan.
\newblock Videocrafter1: Open diffusion models for high-quality video generation.
\newblock 2023.

\bibitem{objaverse}
Matt Deitke, Dustin Schwenk, Jordi Salvador, Luca Weihs, Oscar Michel, Eli VanderBilt, Ludwig Schmidt, Kiana Ehsani, Aniruddha Kembhavi, and Ali Farhadi.
\newblock Objaverse: A universe of annotated 3d objects.
\newblock In {\em CVPR}, 2023.

\bibitem{sora}
{OpenAI}.
\newblock Video generation models as world simulators, 2024.

\bibitem{snapvideo}
Willi Menapace, Aliaksandr Siarohin, Ivan Skorokhodov, Ekaterina Deyneka, Tsai-Shien Chen, Anil Kag, Yuwei Fang, Aleksei Stoliar, Elisa Ricci, Jian Ren, et~al.
\newblock Snap video: Scaled spatiotemporal transformers for text-to-video synthesis.
\newblock In {\em CVPR}, 2024.

\bibitem{vd3d}
Sherwin Bahmani, Ivan Skorokhodov, Aliaksandr Siarohin, Willi Menapace, Guocheng Qian, Michael Vasilkovsky, Hsin-Ying Lee, Chaoyang Wang, Jiaxu Zou, Andrea Tagliasacchi, et~al.
\newblock Vd3d: Taming large video diffusion transformers for 3d camera control.
\newblock 2024.

\bibitem{motionctrl}
Zhouxia Wang, Ziyang Yuan, Xintao Wang, Yaowei Li, Tianshui Chen, Menghan Xia, Ping Luo, and Ying Shan.
\newblock Motionctrl: A unified and flexible motion controller for video generation.
\newblock In {\em SIGGRAPH}, 2024.

\bibitem{cameractrl}
Hao He, Yinghao Xu, Yuwei Guo, Gordon Wetzstein, Bo~Dai, Hongsheng Li, and Ceyuan Yang.
\newblock Cameractrl: Enabling camera control for text-to-video generation.
\newblock 2024.

\bibitem{directavideo}
Shiyuan Yang, Liang Hou, Haibin Huang, Chongyang Ma, Pengfei Wan, Di~Zhang, Xiaodong Chen, and Jing Liao.
\newblock Direct-a-video: Customized video generation with user-directed camera movement and object motion.
\newblock In {\em SIGGRAPH}, 2024.

\bibitem{camco}
Dejia Xu, Weili Nie, Chao Liu, Sifei Liu, Jan Kautz, Zhangyang Wang, and Arash Vahdat.
\newblock Camco: Camera-controllable 3d-consistent image-to-video generation.
\newblock 2024.

\bibitem{wang2024freevs}
Qitai Wang, Lue Fan, Yuqi Wang, Yuntao Chen, and Zhaoxiang Zhang.
\newblock Freevs: Generative view synthesis on free driving trajectory.
\newblock 2024.

\bibitem{webvid10m}
Max Bain, Arsha Nagrani, G{\"u}l Varol, and Andrew Zisserman.
\newblock Frozen in time: A joint video and image encoder for end-to-end retrieval.
\newblock In {\em ICCV}, 2021.

\bibitem{realestate10k}
Richard Tucker and Noah Snavely.
\newblock Stereo magnification: Learning view synthesis using multiplane images.
\newblock In {\em ToG}, 2018.

\bibitem{fan2024instantsplat}
Zhiwen Fan, Kairun Wen, Wenyan Cong, Kevin Wang, Jian Zhang, Xinghao Ding, Danfei Xu, Boris Ivanovic, Marco Pavone, Georgios Pavlakos, Zhangyang Wang, and Yue Wang.
\newblock Instantsplat: Sparse-view gaussian splatting in seconds, 2024.

\bibitem{Jain_2021_ICCV}
Ajay Jain, Matthew Tancik, and Pieter Abbeel.
\newblock Putting nerf on a diet: Semantically consistent few-shot view synthesis.
\newblock In {\em ICCV}, 2021.

\bibitem{chen2024g3r}
Yun Chen, Jingkang Wang, Ze~Yang, Sivabalan Manivasagam, and Raquel Urtasun.
\newblock G3r: Gradient guided generalizable reconstruction.
\newblock In {\em European Conference on Computer Vision}, 2024.

\bibitem{23iccv_tian_mononerf}
Fengrui Tian, Shaoyi Du, and Yueqi Duan.
\newblock {MonoNeRF}: Learning a generalizable dynamic radiance field from monocular videos.
\newblock In {\em ICCV}, 2023.

\bibitem{gntmove2023}
Wenyan Cong, Hanxue Liang, Peihao Wang, Zhiwen Fan, Tianlong Chen, Mukund Varma, Yi~Wang, and Zhangyang Wang.
\newblock Enhancing ne{RF} akin to enhancing {LLM}s: Generalizable ne{RF} transformer with mixture-of-view-experts.
\newblock In {\em ICCV}, 2023.

\bibitem{t2023is}
Mukund~Varma T, Peihao Wang, Xuxi Chen, Tianlong Chen, Subhashini Venugopalan, and Zhangyang Wang.
\newblock Is attention all that ne{RF} needs?
\newblock In {\em ICLR}, 2023.

\bibitem{hong2023lrm}
Yicong Hong, Kai Zhang, Jiuxiang Gu, Sai Bi, Yang Zhou, Difan Liu, Feng Liu, Kalyan Sunkavalli, Trung Bui, and Hao Tan.
\newblock Lrm: Large reconstruction model for single image to 3d.
\newblock {\em ICLR}, 2024.

\bibitem{charatan23pixelsplat}
David Charatan, Sizhe Li, Andrea Tagliasacchi, and Vincent Sitzmann.
\newblock pixelsplat: 3d gaussian splats from image pairs for scalable generalizable 3d reconstruction.
\newblock In {\em CVPR}, 2024.

\bibitem{chen2024mvsplat}
Yuedong Chen, Haofei Xu, Chuanxia Zheng, Bohan Zhuang, Marc Pollefeys, Andreas Geiger, Tat-Jen Cham, and Jianfei Cai.
\newblock Mvsplat: Efficient 3d gaussian splatting from sparse multi-view images.
\newblock {\em ECCV}, 2024.

\bibitem{chen2024mvsplat360}
Yuedong Chen, Chuanxia Zheng, Haofei Xu, Bohan Zhuang, Andrea Vedaldi, Tat-Jen Cham, and Jianfei Cai.
\newblock Mvsplat360: Feed-forward 360 scene synthesis from sparse views.
\newblock 2024.

\bibitem{gslrm2024}
Kai Zhang, Sai Bi, Hao Tan, Yuanbo Xiangli, Nanxuan Zhao, Kalyan Sunkavalli, and Zexiang Xu.
\newblock Gs-lrm: Large reconstruction model for 3d gaussian splatting.
\newblock {\em ECCV}, 2024.

\bibitem{ren2024scube}
Xuanchi Ren, Yifan Lu, Hanxue Liang, Jay~Zhangjie Wu, Huan Ling, Mike Chen, Francis Fidler, Sanja annd~Williams, and Jiahui Huang.
\newblock Scube: Instant large-scale scene reconstruction using voxsplats.
\newblock In {\em NeurIPS}, 2024.

\bibitem{jin2025lvsm}
Haian Jin, Hanwen Jiang, Hao Tan, Kai Zhang, Sai Bi, Tianyuan Zhang, Fujun Luan, Noah Snavely, and Zexiang Xu.
\newblock Lvsm: A large view synthesis model with minimal 3d inductive bias.
\newblock In {\em ICLR}, 2025.

\bibitem{jiang2025rayzer}
Hanwen Jiang, Hao Tan, Peng Wang, Haian Jin, Yue Zhao, Sai Bi, Kai Zhang, Fujun Luan, Kalyan Sunkavalli, Qixing Huang, and Georgios Pavlakos.
\newblock Rayzer: A self-supervised large view synthesis model.
\newblock 2025.

\bibitem{smart2024splatt3r}
Brandon Smart, Chuanxia Zheng, Iro Laina, and Victor~Adrian Prisacariu.
\newblock Splatt3r: Zero-shot gaussian splatting from uncalibrated image pairs.
\newblock 2024.

\bibitem{zhang2025FLARE}
Shangzhan Zhang, Jianyuan Wang, Yinghao Xu, Nan Xue, Christian Rupprecht, Xiaowei Zhou, Yujun Shen, and Gordon Wetzstein.
\newblock Flare: Feed-forward geometry, appearance and camera estimation from uncalibrated sparse views.
\newblock {\em CVPR}, 2025.

\bibitem{ye2024noposplat}
Botao Ye, Sifei Liu, Haofei Xu, Li~Xueting, Marc Pollefeys, Ming-Hsuan Yang, and Peng Songyou.
\newblock No pose, no problem: Surprisingly simple 3d gaussian splats from sparse unposed images.
\newblock {\em ICLR}, 2025.

\bibitem{hong2024pf3plat}
Sunghwan Hong, Jaewoo Jung, Heeseong Shin, Jisang Han, Jiaolong Yang, Chong Luo, and Seungryong Kim.
\newblock Pf3plat: Pose-free feed-forward 3d gaussian splatting.
\newblock {\em ICML}, 2025.

\bibitem{zhao2024pgdvs}
Xiaoming Zhao, Alex Colburn, Fangchang Ma, Miguel Ángel Bautista, Joshua~M. Susskind, and Alexander~G. Schwing.
\newblock {Pseudo-Generalized Dynamic View Synthesis from a Video}.
\newblock In {\em ICLR}, 2024.

\bibitem{zhang2024monst3r}
Junyi Zhang, Charles Herrmann, Junhwa Hur, Varun Jampani, Trevor Darrell, Forrester Cole, Deqing Sun, and Ming-Hsuan Yang.
\newblock Monst3r: A simple approach for estimating geometry in the presence of motion.
\newblock {\em ICLR}, 2025.

\bibitem{Wang_2024_CVPR}
Shuzhe Wang, Vincent Leroy, Yohann Cabon, Boris Chidlovskii, and Jerome Revaud.
\newblock Dust3r: Geometric 3d vision made easy.
\newblock In {\em CVPR}, 2024.

\bibitem{mast3r_arxiv24}
Vincent Leroy, Yohann Cabon, and Jerome Revaud.
\newblock Grounding image matching in 3d with mast3r, 2024.

\bibitem{ren2024l4gm}
Jiawei Ren, Kevin Xie, Ashkan Mirzaei, Hanxue Liang, Xiaohui Zeng, Karsten Kreis, Ziwei Liu, Antonio Torralba, Sanja Fidler, Seung~Wook Kim, and Huan Ling.
\newblock L4gm: Large 4d gaussian reconstruction model.
\newblock In {\em NeurIPS}, 2024.

\bibitem{liang2024btimer}
Hanxue Liang, Jiawei Ren, Ashkan Mirzaei, Antonio Torralba, Ziwei Liu, Igor Gilitschenski, Sanja Fidler, Cengiz Oztireli, Huan Ling, Zan Gojcic, and Jiahui Huang.
\newblock Feed-forward bullet-time reconstruction of dynamic scenes from monocular videos.
\newblock 2024.

\bibitem{Peebles2022DiT}
William Peebles and Saining Xie.
\newblock Scalable diffusion models with transformers.
\newblock 2022.

\bibitem{cosmos}
Niket Agarwal, Arslan Ali, Maciej Bala, Yogesh Balaji, Erik Barker, Tiffany Cai, Prithvijit Chattopadhyay, Yongxin Chen, Yin Cui, Yifan Ding, Daniel Dworakowski, Jiaojiao Fan, Michele Fenzi, Francesco Ferroni, Sanja Fidler, Dieter Fox, Songwei Ge, Yunhao Ge, Jinwei Gu, Siddharth Gururani, Ethan He, Jiahui Huang, Jacob Huffman, Pooya Jannaty, Jingyi Jin, Seung~Wook Kim, Gergely Klár, Grace Lam, Shiyi Lan, Laura Leal-Taixe, Anqi Li, Zhaoshuo Li, Chen-Hsuan Lin, Tsung-Yi Lin, Huan Ling, Ming-Yu Liu, Xian Liu, Alice Luo, Qianli Ma, Hanzi Mao, Kaichun Mo, Arsalan Mousavian, Seungjun Nah, Sriharsha Niverty, David Page, Despoina Paschalidou, Zeeshan Patel, Lindsey Pavao, Morteza Ramezanali, Fitsum Reda, Xiaowei Ren, Vasanth Rao~Naik Sabavat, Ed~Schmerling, Stella Shi, Bartosz Stefaniak, Shitao Tang, Lyne Tchapmi, Przemek Tredak, Wei-Cheng Tseng, Jibin Varghese, Hao Wang, Haoxiang Wang, Heng Wang, Ting-Chun Wang, Fangyin Wei, Xinyue Wei, Jay~Zhangjie Wu, Jiashu Xu, Wei Yang, Lin Yen-Chen, Xiaohui Zeng, Yu~Zeng, Jing
  Zhang, Qinsheng Zhang, Yuxuan Zhang, Qingqing Zhao, and Artur Zolkowski.
\newblock Cosmos world foundation model platform for physical ai, 2025.

\bibitem{sv4d-v2}
Chun-Han Yao, Yiming Xie, Vikram Voleti, Huaizu Jiang, and Varun Jampani.
\newblock Sv4d 2.0: Enhancing spatio-temporal consistency in multi-view video diffusion for high-quality 4d generation, 2025.

\bibitem{flexattention}
Juechu Dong, Boyuan Feng, Driss Guessous, Yanbo Liang, and Horace He.
\newblock Flex attention: A programming model for generating optimized attention kernels, 2024.

\bibitem{liu2022flow}
Xingchao Liu, Chengyue Gong, and Qiang Liu.
\newblock Flow straight and fast: Learning to generate and transfer data with rectified flow.
\newblock 2022.

\bibitem{kubric}
Klaus Greff, Francois Belletti, Lucas Beyer, Carl Doersch, Yilun Du, Daniel Duckworth, David~J Fleet, Dan Gnanapragasam, Florian Golemo, Charles Herrmann, Thomas Kipf, Abhijit Kundu, Dmitry Lagun, Issam Laradji, Hsueh-Ti~(Derek) Liu, Henning Meyer, Yishu Miao, Derek Nowrouzezahrai, Cengiz Oztireli, Etienne Pot, Noha Radwan, Daniel Rebain, Sara Sabour, Mehdi S.~M. Sajjadi, Matan Sela, Vincent Sitzmann, Austin Stone, Deqing Sun, Suhani Vora, Ziyu Wang, Tianhao Wu, Kwang~Moo Yi, Fangcheng Zhong, and Andrea Tagliasacchi.
\newblock Kubric: a scalable dataset generator.
\newblock 2022.

\bibitem{yoon2020novel}
Jae~Shin Yoon, Kihwan Kim, Orazio Gallo, Hyun~Soo Park, and Jan Kautz.
\newblock Novel view synthesis of dynamic scenes with globally coherent depths from a monocular camera.
\newblock In {\em CVPR}, 2020.

\bibitem{zhang2018perceptual}
Richard Zhang, Phillip Isola, Alexei~A Efros, Eli Shechtman, and Oliver Wang.
\newblock The unreasonable effectiveness of deep features as a perceptual metric.
\newblock In {\em CVPR}, 2018.

\bibitem{ling2024dl3dv}
Lu~Ling, Yichen Sheng, Zhi Tu, Wentian Zhao, Cheng Xin, Kun Wan, Lantao Yu, Qianyu Guo, Zixun Yu, Yawen Lu, et~al.
\newblock Dl3dv-10k: A large-scale scene dataset for deep learning-based 3d vision.
\newblock In {\em CVPR}, 2024.

\bibitem{yu2023mvimgnet}
Xianggang Yu, Mutian Xu, Yidan Zhang, Haolin Liu, Chongjie Ye, Yushuang Wu, Zizheng Yan, Tianyou Liang, Guanying Chen, Shuguang Cui, and Xiaoguang Han.
\newblock Mvimgnet: A large-scale dataset of multi-view images.
\newblock In {\em CVPR}, 2023.

\bibitem{infinite_nature_2020}
Andrew Liu, Richard Tucker, Varun Jampani, Ameesh Makadia, Noah Snavely, and Angjoo Kanazawa.
\newblock Infinite nature: Perpetual view generation of natural scenes from a single image.
\newblock In {\em ICCV}, 2021.

\bibitem{vbench++}
Ziqi Huang, Fan Zhang, Xiaojie Xu, Yinan He, Jiashuo Yu, Ziyue Dong, Qianli Ma, Nattapol Chanpaisit, Chenyang Si, Yuming Jiang, Yaohui Wang, Xinyuan Chen, Ying-Cong Chen, Limin Wang, Dahua Lin, Yu~Qiao, and Ziwei Liu.
\newblock Vbench++: Comprehensive and versatile benchmark suite for video generative models.
\newblock 2024.

\bibitem{met3r}
Mohammad Asim, Christopher Wewer, Thomas Wimmer, Bernt Schiele, and Jan~Eric Lenssen.
\newblock Met3r: Measuring multi-view consistency in generated images.
\newblock In {\em CVPR}, 2024.

\bibitem{Knapitsch2017tandt}
Arno Knapitsch, Jaesik Park, Qian-Yi Zhou, and Vladlen Koltun.
\newblock Tanks and temples: Benchmarking large-scale scene reconstruction.
\newblock {\em ToG}, 2017.

\bibitem{mildenhall2019llff}
Ben Mildenhall, Pratul~P. Srinivasan, Rodrigo Ortiz-Cayon, Nima~Khademi Kalantari, Ravi Ramamoorthi, Ren Ng, and Abhishek Kar.
\newblock Local light field fusion: Practical view synthesis with prescriptive sampling guidelines.
\newblock {\em ToG}, 2019.

\bibitem{li2022neural3dvideosynthesis}
Tianye Li, Mira Slavcheva, Michael Zollhoefer, Simon Green, Christoph Lassner, Changil Kim, Tanner Schmidt, Steven Lovegrove, Michael Goesele, Richard Newcombe, and Zhaoyang Lv.
\newblock Neural 3d video synthesis from multi-view video.
\newblock In {\em CVPR}, 2022.

\end{thebibliography}

\end{document}